%% file: main.tex
\documentclass[10pt,twocolumn,letterpaper]{article}

\usepackage[pagenumbers]{cvpr} % To force page numbers, e.g. for an arXiv version

\input{preamble}

\usepackage{multirow} 
\usepackage{array}
\usepackage{pifont}

\definecolor{cvprblue}{rgb}{0.21,0.49,0.74}
\usepackage[pagebackref,breaklinks,colorlinks,citecolor=cvprblue]{hyperref}

\def\paperID{13885} % *** Enter the Paper ID here
\def\confName{CVPR}
\def\confYear{2025}

\definecolor{mycolor}{RGB}{147,112,219}

\title{LinGen: Towards High-Resolution Minute-Length Text-to-Video Generation with Linear Computational Complexity}

\author{Hongjie Wang$^{1,2}$, Chih-Yao Ma$^2$, Yen-Cheng Liu$^2$, Ji Hou$^2$, Tao Xu$^2$, Jialiang Wang$^2$, Felix Juefei-Xu$^2$, \\
Yaqiao Luo$^2$, Peizhao Zhang$^2$, Tingbo Hou$^2$, Peter Vajda$^2$, Niraj K. Jha$^1$, Xiaoliang Dai$^2$\\
$^1$Princeton University, $^2$Meta
}

\begin{document}
\maketitle
\input{sec/0_abstract}    
\input{sec/1_intro}

\input{sec/2_related_work}
\input{sec/3_methodology}
\input{sec/4_experiments}

\input{sec/5_conclusion}

{
    \small
    \bibliographystyle{ieeenat_fullname}
    \bibliography{main}
}

% % WARNING: do not forget to delete the supplementary pages from your submission 

\input{sec/X_suppl}

\end{document}

%% file: preamble.tex
\usepackage[dvipsnames,table]{xcolor}

\definecolor{americanrose}{rgb}{1.0, 0.01, 0.24}

%% file: sec/0_abstract.tex
\begin{abstract}

Text-to-video generation enhances content creation but is highly computationally intensive: 
% Although Diffusion Transformers (DiTs) with billions of parameters can generate high-quality videos, 
The computational cost of Diffusion Transformers (DiTs) scales quadratically in the number of pixels.  This makes minute-length video generation extremely expensive, limiting most existing models to generating videos of only 10-20 seconds length.
% Although Diffusion Transformers (DiTs) have been scaled to tens of billion parameters to generate high-quality videos, their computational cost scales quadratically with the number of pixels.
% (that depends on resolution and length of generated videos). 
% It makes minute-length text-to-video generation extremely expensive, restricting most existing models to generating videos of only 10-20 seconds.
% (most of existing works is only able to generation 10-20s videos) \textbf{mention the impractical thing of quadratic. What does linear complexity enable?} 
We propose a \textbf{Lin}ear-complexity text-to-video \textbf{Gen}eration (\textbf{LinGen}) framework whose cost scales linearly in the number of pixels. For the first time, LinGen enables high-resolution minute-length video generation on a single GPU without compromising quality. It replaces the computationally-dominant and quadratic-complexity block, self-attention, with a linear-complexity block called 
% \felix{shall we mention what MATE stands for first? Hongjie: The name comes from its two key components: MAMBA and TESA. And the word ``mate" itself also means a ``couple"}
\textbf{MATE}, which consists of an \textbf{MA}-branch and a \textbf{TE}-branch. The MA-branch targets short-to-long-range correlations, combining a bidirectional Mamba2 block with our token rearrangement method, Rotary Major Scan, and our review tokens developed for long video generation. The TE-branch is a novel TEmporal Swin Attention block that focuses on temporal correlations between adjacent tokens and medium-range tokens. 
% Native Mamba leads to significant inconsistency in generated videos due to the adjacency preservation issue and long-range correlation decay. 
The MATE block addresses the adjacency preservation issue of Mamba and improves the consistency of generated videos significantly. 
% We further implement review tokens to enhance long-range correlation for long video generation. 
Experimental results show that  LinGen outperforms DiT (with a 75.6\% win rate) in video quality with up to 15$\times$ (11.5$\times$) FLOPs (latency) reduction.  Furthermore, both automatic metrics and human evaluation demonstrate our LinGen-4B yields comparable video quality to state-of-the-art models (with a 50.5\%, 52.1\%, 49.1\% win rate with respect to Gen-3, LumaLabs, and Kling, respectively). This paves the way to hour-length movie generation and real-time interactive video generation. 
% \kevin{which version of Kling?}
We provide 68s video generation results and more examples in our project website: \href{https://lineargen.github.io/}{https://lineargen.github.io/}.
% \kevin{In general, I think the abstract reads good but is a bit long. It's fine if we are within page limit. Though, some of the content regarding the MATE and its branches are pretty similar of what mentioned in the intro.} 

% Experimental results indicate that our LinGen-4B model outperforms the DiT-4B baseline with global self-attention and generates videos that have comparable quality to state-of-the-art video generation models, 
% while achieving linear complexity and up to 7.5$\times$ latency reduction when generating 512p 68s videos at 16fps \xiaoliang{didn't we just claim DiT cannot generate 512p 68s videos}. This paves the way to hour-length movie generation and real-time interactive video generation.
% \textbf{why don't we need to compromise quality?} 
% Please check the visual examples of generated videos in our supplmentary material.
\end{abstract}

%% file: sec/1_intro.tex
\vspace{-1em}
\section{Introduction}
\label{sec:intro}

Diffusion Models (DMs)~\cite{ho2020ddpm,sohl2015diffusion} have exhibited superior performance on various generative tasks, including image generation~\cite{rombach2022ldm,dai2023emu,podell2023sdxl, saharia2022imagen}, % super-resolution~\cite{li2022srdiff,gao2023idm}, 
image editing~\cite{sheynin2024emuedit,yang2023paint,kawar2023imagic,couairon2022diffedit},  3D shape generation~\cite{vahdat2022lion,luo2021diffusion3d}, 
% audio generation~\cite{kong2020diffwave,huang2023makeanaudio}, 
% text generation~\cite{li2022diffusionlm,gong2022diffuseq,loudiscrete}, \xiaoliang{maybe remove the text generation, doesn't seem relevant to us} 
and video generation~\cite{sora2024,polyak2024moviegen,opensora,girdhar2023emuvideo}. Among them, high-resolution text-to-video generation is widely regarded as one of the most challenging tasks due to two key factors: (1) the immense complexity of predicting the values of hundreds of millions of pixels and (2) the human eye’s acute sensitivity to inconsistencies across frames. Sora~\cite{sora2024} and Movie Gen~\cite{polyak2024moviegen} achieve highly consistent video generation by scaling Diffusion Transformers (DiTs)~\cite{peebles2023dit} to tens of billions of parameters. However, the computational cost of DiTs scales quadratically in the resolution and length of generated videos, making it extremely expensive to generate long videos and limiting the raw video length of most existing models to 10-20 seconds.% with high fidelity. 
% \xiaoliang{maybe mention the typical video generation length}
% \xiaoliang{the logic doesn't sound correct: why do personal users need minute-length generation?} which makes the technology inaccessible to small companies and personal users. 

\begin{figure*}[t]
  \centering
   \includegraphics[width=\linewidth]{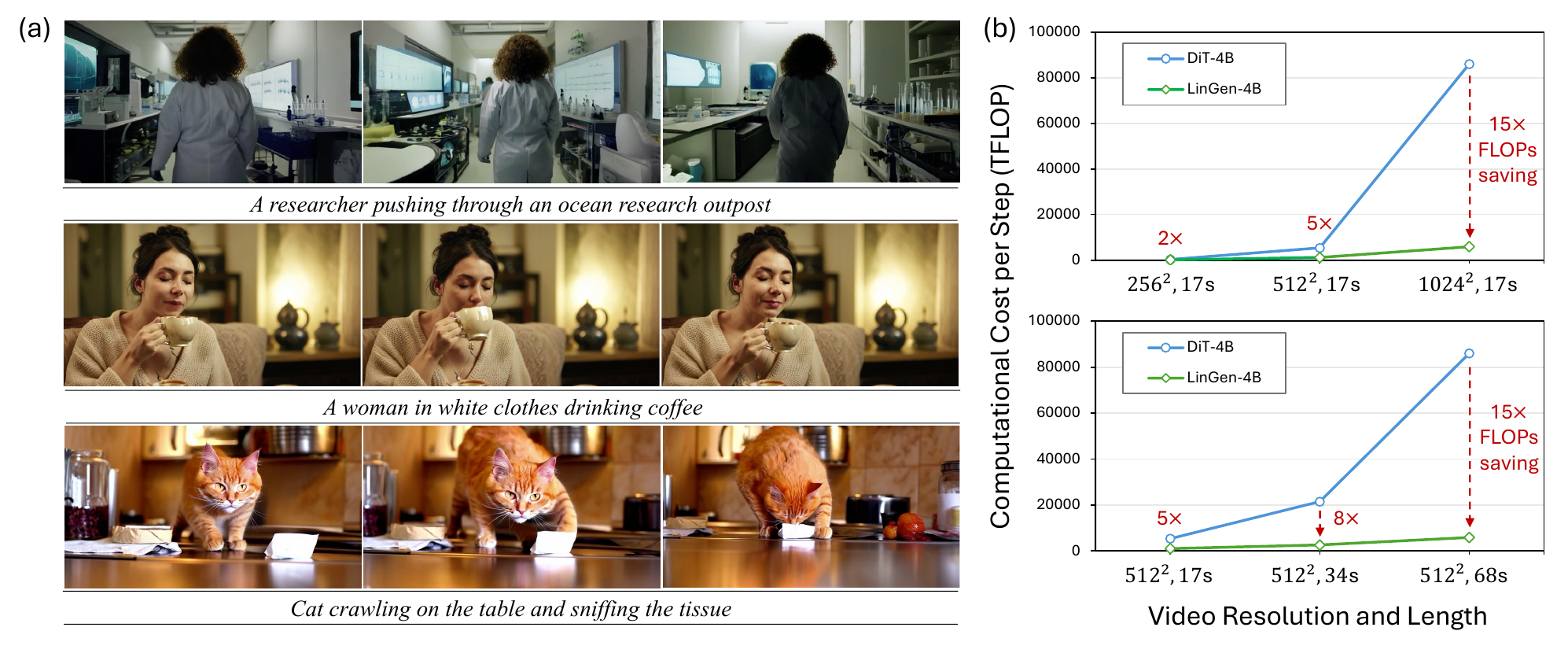}
   \vspace{-2.5em}
   \caption{\textbf{LinGen generates photorealistic high-resolution long videos with linear computational complexity.}
   % \yc{change scaling law?} 
   (a) High-quality videos generated using our LinGen model. (b) The computational cost scaling curves across different video resolutions and lengths. LinGen achieves 15$\times$ speed-up compared to the standard DiT when generating 68s-length videos at 512p resolution.
   % \kevin{this seems to be a figure best fit for 1st page?}
   }
   \vspace{-1.5em}
   \label{fig:teaser}
\end{figure*}

Numerous existing studies have focused on improving the efficiency of video generation. This can be categorized into two approaches: (1) sampling distillation~\cite{li2024t2vturbo, wang2023videolcm}, which reduces the number of sampling steps, and (2) efficient architectural designs that lower the computational cost of each sampling step, which includes factorized attention~\cite{chen2023videocrafter1,wang2023modelscope} and State Space Models (SSMs)~\cite{gao2024matten,mo2024dim}. However, they either retain quadratic complexity or are restricted to generating low-resolution, short videos. It is challenging to perform high-resolution long video generation solely based on the linear-complexity SSMs like Mamba~\cite{gu2023mamba}, due to its \textbf{\textit{adjacency preservation issue}}~\cite{gao2024matten}.
% Nevertheless, the computational cost of both factorized attention and the SSM-attention hybrid architecture still scales quadratically in the resolution and length of generated videos. DiM~\cite{mo2024dim} achieves linear-complexity generation by replacing all attention layers with an efficient variant of SSM, Mamba~\cite{gu2023mamba}, but it performs video generation at a small scale without text prompt. Simply scaling up this architecture leads to significant inconsistency in generated videos due to the \textbf{\textit{adjacency preservation issue}} of Mamba~\cite{gao2024matten}. 
Mamba was originally designed for language tasks, where the inputs are natively sequences. When it is adapted to the vision modality, rearranging 2D (images) or 3D (videos) tensors into a 1D sequence becomes a necessity. This rearrangement causes spatially and temporally adjacent tokens to become distant in the sequence. This significantly hurts the quality of generated images and videos~\cite{hu2024zigma} due to the inherent decay when Mamba calculates long-range correlations~\cite{gu2023mamba}. Although more sophisticated rearrangement methods~\cite{hu2024zigma,he2024mambaad,ren2024mambacsr} could alleviate this issue, they can hardly ensure consistency across frames when scaled to high-resolution long video generation.
% \kevin{do we have experiment to back it up? If so, let's say see Sec. X for results and analysis.}

To address the above challenge, we propose a \textbf{Linear-complexity text-to-video Generation (LinGen)} framework that scales linearly in the number of pixels in generated videos. To the best of our knowledge, LinGen is \textbf{the first to enable photorealistic high-resolution minute-length video generation at a high frame rate on a single GPU without video extension, super-resolution, or compromising quality}. It not only addresses the aforementioned adjacency preservation issue (see Supplementary Sec.~\ref{app:adjacency}), but also comprehensively enhances the short-, medium-, and long-range correlations while maintaining linear complexity.
% \kevin{to claim that we can "comprehensively enhances the short-, medium-, and long-range correlations" seems to be a strong statement. What experiments do we have to back it up?}
LinGen replaces the self-attention layers in DiTs with our proposed linear-complexity MATE blocks. 
Each MATE block is composed of an MA-branch and a TE-branch. The MA-branch consists of a bidirectional Mamba2~\cite{dao2024mamba2} (a transformer-format SSM variant) block equipped with our proposed Rotary-Major Scan (RMS) and review tokens. RMS rearranges 3D token tensors in the latent space before they enter the bidirectional Mamba2 block, enhancing short-range correlations. To alleviate the inherent long-range correlation decay of SSMs, review tokens provide an overview of the processed token sequences to the hidden state of Mamba2 blocks at the start of sequence processing, to calibrate long-range correlations. The TE-branch is a novel TEmporal Swin Attention (TESA) block. It computes correlations among short-range spatially adjacent and medium-range temporally adjacent tokens, focusing on addressing the adjacency preservation issue and improving video consistency.
% To further enhancing adjacent correlations, 3D token tensors in the latent space are rearranged into sequences using our proposed Rotary-Major Scan (RMS) method before entering the bidirectional Mamba2 blocks. 
% Together, TESA and RMS address the adjacency preservation challenge by accurately capturing adjacent correlations, resulting in significant improvements in consistency and fidelity of generated videos. 
% Furthermore, to account for the inherent long-range correlation decay of SSMs, we introduce review tokens to calibrate long-range correlations for generating long videos. They provide an overview of the processed token sequences to the hidden state of Mamba2 blocks at the start of sequence processing.
% \yc{may briefly introduce the review tokens} 
Note that LinGen is orthogonal to sampling distillation and can potentially be combined with it to further boost its efficiency. Our contributions can be summarized as follows.
% \kevin{most of the content above is for "introduction of the proposed method", should we reduce the length a bit and instead mention highlights of the experimental results and speed up?}

% \vspace{-0.5em}
\begin{itemize}
    \item We propose LinGen, a text-to-video generation framework that enables photorealistic minute-length video generation with linear computational complexity.
    \item To comprehensively cover short-, medium-, and long-range correlations, we compose our proposed self-attention replacement block, MATE, with an MA-branch, including a bidirectional Mamba2 block equipped with our RMS and review tokens, and a TE-branch that includes a novel TESA block.
    % \item For better long video generation, we further introduce review tokens to calibrate long-range correlations and compensate for the inherent correlation decay in SSMs.
    \item We establish the superiority of the proposed LinGen framework by comparing it to our self-attention baseline, DiT-4B, and other existing video generation models via human evaluations and automatic evaluation metrics. Experimental results indicate LinGen generates photorealistic high-quality videos while achieving linear scaling and up to 15$\times$ speed-up when generating minute-length videos at 16 fps (see Fig.~\ref{fig:teaser}).
\end{itemize}

%% file: sec/2_related_work.tex
\section{Related Work}
\label{sec:related}

\begin{figure*}[t]
  \centering
   \includegraphics[width=0.95\linewidth]{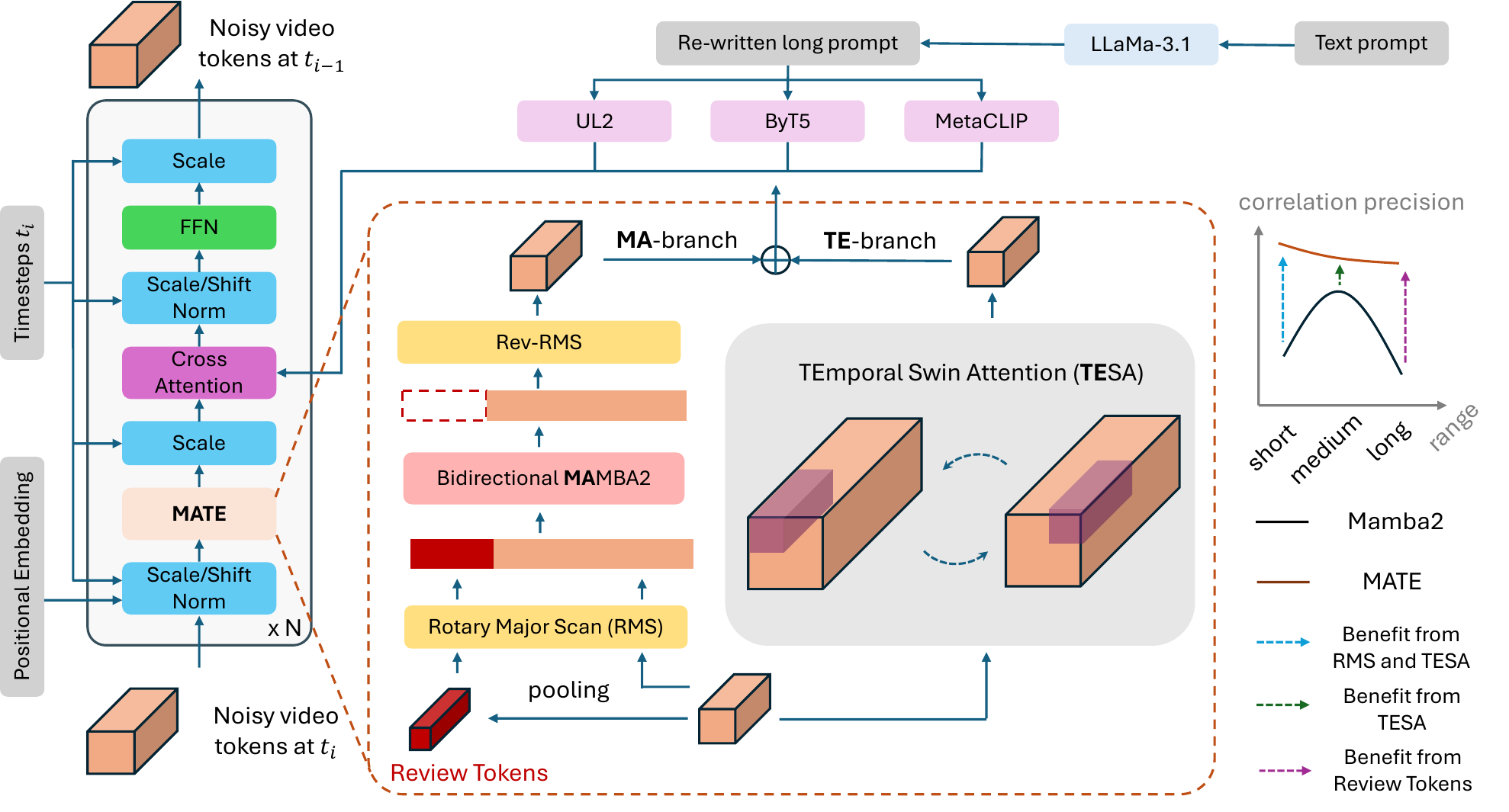}
   \vspace{-1em}
   \caption{\textbf{Overview of the LinGen denoising module.} 
   % Text prompts are rewritten by Llama-3.1 and projected to text embeddings via three text encoders: UL2, ByT5, and MetaCLIP. 
   LinGen replaces self-attention layers with a MATE block, which inherits linear complexity from its two branches: \textbf{MA-branch} and \textbf{TE-branch}. The \textbf{MA-branch} consists of a bidirectional Mamba2 block, RMS, and review tokens to cover short-to-long-range correlations. The \textbf{TE-branch} is a TEmporal Swin Attention block that addresses the adjacency preservation issue and improves the consistency of generated videos significantly.
   % improving the short-range correlation precision significantly. To assist long-range correlation calculation in long video generation, MATE incorporates review tokens, which provide an overview of previously generated frames.
   }
   \vspace{-1.5em}
   \label{fig:overview}
\end{figure*}

\noindent\textbf{High-Quality Video Generation.} 
% Video generation is widely recognized as one of the most challenging tasks in generative modeling. 
Sora~\cite{sora2024} was the first work to successfully produce high-resolution videos with exceptional consistency. It learns an encoded latent space and deploys a large-scale DiT embedded in it. Runway Gen3~\cite{runwaygen32024}, LumaLabs~\cite{lumalabs2024}, and Kling~\cite{klingai2024} are subsequent works capable of generating highly consistent, high-resolution videos with high frame rates. 
% However, all three are proprietary commercial models, with no implementation details publicly disclosed. 
MovieGen~\cite{polyak2024moviegen} generates photorealistic and highly consistent videos with all implementation details revealed. However, it scales the DiT to 30 billion parameters. Its quadratic complexity makes generating minute-length videos very difficult. 
Several open-source models~\cite{opensora,wang2023modelscope,chen2024videocrafter2} also aim to generate high-quality videos. However, the quality of their outputs still notably lags behind that of the aforementioned models. An alternative to DMs for video generation is the use of transformer-based language models, which auto-regressively generate video tokens~\cite{kondratyuk2023videopoet,yu2023magvit,nash2022transframer,villegas2022phenaki,yan2021videogpt}. While these models are well-suited to multimodal conditioning tasks, the quality of their generated videos generally falls short of that achieved by DM-based models. 

\noindent\textbf{Efficient Video Generation.} The high computational cost of DM-based video generation has prompted various research efforts to address this challenge. Most of them are inspired by efficient DM-based image generation works~\cite{luo2023lcm,meng2023distillation,kim2023architectural,wang2024atedm} and can be divided into two types: (1) \textbf{Sampling distillation} to reduce the required number of sampling steps to generate high-quality videos. VideoLCM~\cite{wang2023videolcm} uses Consistency Distillation~\cite{song2023consistency} to enable satisfactory video generation in four steps. T2V-Turbo~\cite{li2024t2vturbo} integrates reward feedback into the distillation process to further improve video quality. (2) \textbf{Efficient denoising architecture design} to reduce the cost of each sampling step. Many existing works~\cite{chen2023videocrafter1,wang2023modelscope,singer2022makeavideo,ho2022imagenvideo,wang2023lavie} employ factorized spatial and temporal attention to reduce the computational cost of calculating global attention across the entire 3D video token tensor. They still maintain quadratic complexity. Matten~\cite{gao2024matten} and DiM~\cite{mo2024dim} replace some self-attention layers with bidirectional Mamba blocks. However, they either need to maintain some global self-attention layers (thus have quadratic complexity) or can only generate low-resolution short videos. On the contrary, LinGen solves the adjacency preservation issue well and manages to generate high-quality minute-length videos.

\noindent\textbf{Minute-Length Video Generation.} Some existing works~\cite{wang2024loong,xie2024progressive} have conducted early explorations into generating minute-length videos. However, their generated videos have various limitations, including low frame rates, low resolution, and reduced quality due to the extension-based generation pattern. 

%% file: sec/3_methodology.tex
\section{Methodology}
\label{sec:method}

The computational cost of self-attention scales quadratically with the number of tokens in the sequence, creating a bottleneck for DiT-based video generative models due to the extensive length of the encoded video token sequence~\cite{lu2023vdt,wang2023lavie}. Such a quadratic complexity makes generating high-resolution minute-length videos extremely expensive. Therefore, we propose LinGen, a text-to-video generation framework that produces photorealistic videos with linear complexity, enabling high-resolution minute-length video generation at a low cost.

\subsection{Overview}

LinGen uses a Temporal AutoEncoder design that is similar to a prior work~\cite{polyak2024moviegen}. In the latent space, LinGen denoises tokens using Flow Matching~\cite{lipman2022flow} and the linear-quadratic t-schedule~\cite{polyak2024moviegen}. The denoising module of LinGen is shown in Fig.~\ref{fig:overview}. We provide more implementation details in the Supplementary Material (Supp. Mat.) section. 
The cross-attention layer conditions on text embeddings projected by three encoders: UL2~\cite{tay2022ul2}, ByT5~\cite{xue2022byt5}, and MetaCLIP~\cite{xu2023metaclip}. They take long prompts re-written by LLaMa-3.1~\cite{dubey2024llama3} as input. Most importantly, LinGen replaces the self-attention layer of vanilla DiTs with our proposed MATE block, achieving linear computational complexity. MATE is composed of two branches: \textbf{MA-branch} and \textbf{TE-branch}. The \textbf{MA-branch} incorporates a bidirectional Mamba2 block, RMS, to enhance short-range correlations, and review tokens to calibrate long-range correlations (see Sec.~\ref{sec:mabranch}). The \textbf{TE-branch} is a novel TESA block, focusing on correlations among short-range spatially adjacent and medium-range temporally adjacent tokens (see Sec.~\ref{sec:tebranch}). As opposed to Mamba, MATE addresses its adjacency preservation issue and comprehensively enhances short-, medium-, and long-range correlations while maintaining linear complexity in the number of tokens. We describe these components in detail in the following sections and introduce our training recipe in Sec.~\ref{sec:recipe}.
% \yc{the above paragraph may needs to be re-organized and highlight the major proposed components. The current paragraph is more like plain description on architecture/implementation details. One way is to reduce some on detailed description on base model to experiments or supplementary.}

\subsection{MA-Branch: Targets Short-to-Long Range}
\label{sec:mabranch}

% \yc{I feel MATE block should be emphasized more and explicitly. Now the flow is like Mamba $\rightarrow$ scan $\rightarrow$ window attention.  The subtitle of 3.2 -3.4 is module proposed by other works, and this flow makes the method section mostly from other works.
% I would suggest to explicitly re-organize a section and make the sub-title as MATE block, and then mentioning the MATE block includes mamba2 and other modules. }

\begin{figure}[t]
  \centering
   \includegraphics[width=\linewidth]{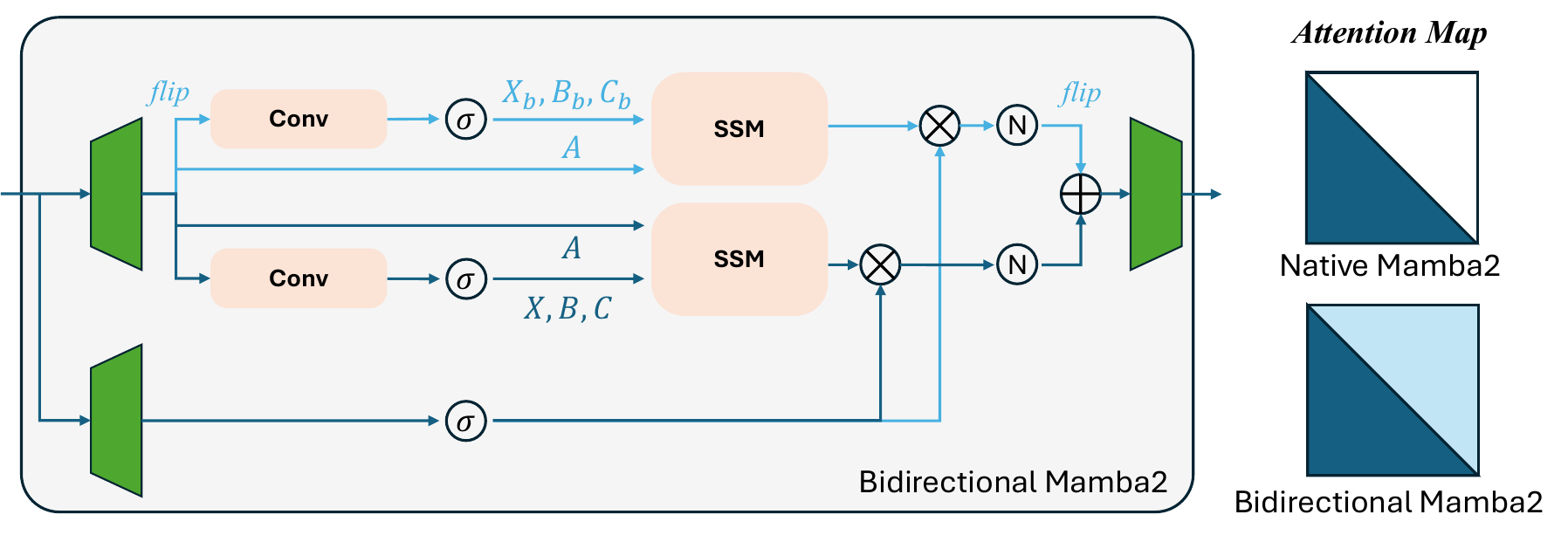}
   \vspace{-2em}
   \caption{\textbf{The bidirectional Mamba2 module. }
   Native Mamba2 only generates the lower triangular part of the attention map due to its causal characteristic. Thus, we deploy bidirectional Mamba2 to obtain the complete attention map for vision tasks.}
   \label{fig:bimamba}
   \vspace{-1.5em}
\end{figure}

\noindent\textbf{Bidirectional Mamba2.} Mamba2~\cite{dao2024mamba2} unifies SSMs and masked efficient attention by proposing a special SSM with an attention format (\ie, Structured State Space Duality). Compared to Mamba, Mamba2 is more hardware-friendly. Thus, we deploy the bidirectional version of Mamba2 in LinGen to obtain the complete correlation map, as shown in Fig.~\ref{fig:bimamba}. The number of FLoating Point Operations (FLOPs) of this block is given by

\vspace{-1.5em}
\begin{equation}
    C_{\text{bimamba}} = (6+\frac{2}{d_h})ENd^2 + 4Nd_sd + O(Nd),
\end{equation}
\vspace{-1em}

\noindent
where $E$ is the expansion factor, $d$ is the dimension of token embedding vectors, $N$ is the number of tokens, $d_s$ is the hidden state size, and $d_h$ is the head dimension of Mamba2, whose default value is 64. We provide the complete expression for $C_{\text{bimamba}}$ in Supp. Mat. This format shows that $C_{\text{bimamba}}$ scales linearly in $N$. The linear complexity of Mamba and Mamba2 makes them highly suitable for video generation, where latent space sequences often contain tens or even hundreds of thousands of tokens. However, videos generated by the native Mamba model exhibit high inconsistency, primarily due to the adjacency preservation issue when rearranging 3D tensor tokens into a sequence~\cite{hu2024zigma,gao2024matten}. Previous works have attempted to address this problem by mixing Mamba layers with global attention layers~\cite{gao2024matten}, thus compromising linear complexity. On the contrary, we equip Mamba2 with RMS and review tokens to build the MA-branch and develop the TE-branch with TESA, enhancing control over continuous spatial and temporal neighbors and calibrating long-range correlations while maintaining linear complexity. 
% We further incorporate review tokens to improve the performance of LinGen when generating long videos.
% \yc{I feel the details of this paragraph can be reduced a bit since most contents in this paragraph are about Mamba2 details not the thing we are proposing. }

% \subsection{Rotary-Major Scan}
% \label{sec:rms}

\begin{figure}[t]
  \centering
   \includegraphics[width=\linewidth]{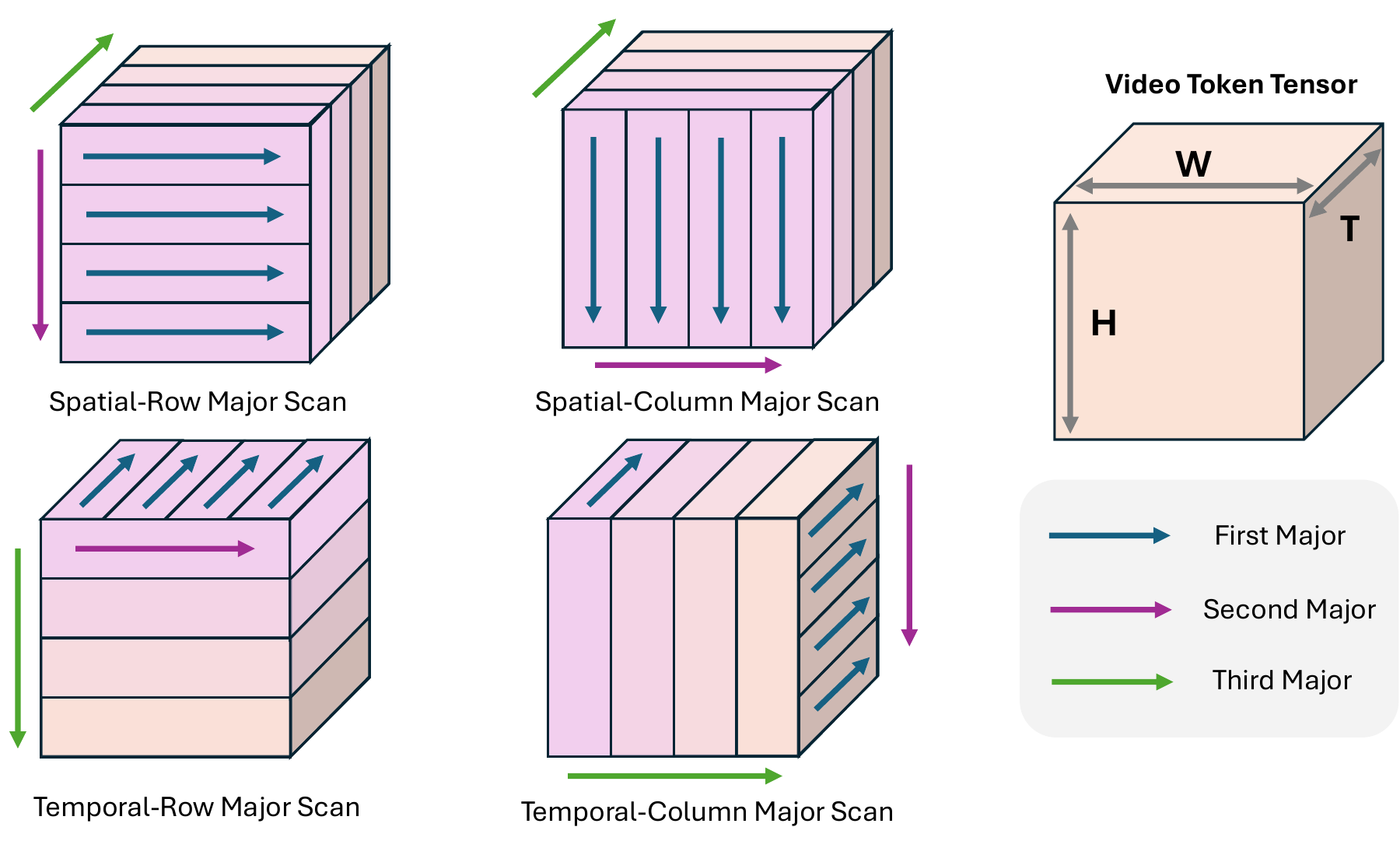}
   \vspace{-2em}
   \caption{\textbf{Rotary-Major Scan (RMS).} We apply different scan schedules across layers to preserve adjacency along various dimensions. Note that scan is bidirectional in practice, but for clarity, only one direction is illustrated for each scan schedule.}
   \vspace{-1.5em}
   \label{fig:rms}
\end{figure}

\noindent\textbf{Rotary-Major Scan.} 
% The adjacency preservation issue of Mamba was originally discovered on the text-to-image generation task~\cite{hu2024zigma}. 
Assume the token tensor shape in the latent space is $H\times W$. Adjacent tokens in the same column are separated at a distance of $H$ in the default row-major scan. Taking into account that Mamba-calculated correlation precision decays as the distance increases, the failure of adjacency preservation leads to distortion in generated images. Zigzag scan~\cite{hu2024zigma} was proposed to alleviate this issue, but it causes significant latency increment when rearranging huge 3D tensors for video generation (see Sec.~\ref{sec:ablation}).
% and is not powerful enough to completely address the adjacency preservation issue (see Sec.~\ref{sec:ablation}).

Thus, we propose RMS, which causes negligible extra latency when targeting large 3D video token tensors. It rearranges the 3D tensor that represents the latent video into a 1D sequence in four different ways in different layers, including spatial-row major, spatial-column major, temporal-row major, and temporal-column major, as shown in Fig.~\ref{fig:rms}. We employ these different scan methods in different layers in an alternating fashion. Assuming the token tensor shape in the latent space is $T\times H\times W$, the index of token $T[t][y][x]$ in the re-arranged 1D sequence in the $l$-th layer is given by

\vspace{-2em}
{\small
\[
n_l = 
\begin{cases} 
      t \cdot (H \cdot W) + y \cdot W + x, & \text{if } l \mod 4 = 0 \\
      t \cdot (H \cdot W) + x \cdot H + y, & \text{if  } l \mod 4 = 1 \\
      y \cdot (T \cdot W) + x \cdot T + t, & \text{if } l \mod 4 = 2 \\
      x \cdot (T \cdot H) + y \cdot T + t, & \text{if } l \mod 4 = 3 
\end{cases}
\]
}
\vspace{-1em}

\noindent
Note that the scan in each layer is bidirectional; hence, a flipped sequence $n_{l,flip}=T\cdot H\cdot W - n_l$ always exists simultaneously. RMS can be implemented with just a few lines of code to reshape the token tensor, making it highly hardware-friendly for processing large tensors. Ablation experiments (see Sec.~\ref{sec:ablation}) show that RMS achieves similar performance to the Zigzag scan in video generation while significantly reducing additional latency.

% \subsection{Review Tokens}
% \label{sec:reviewtokens}

\noindent\textbf{Review Tokens.} To enhance the overall understanding of generated videos and improve text-video alignment in long video generation, we add review tokens when processing extremely long sequences. Specifically, we append an average-pooled version of the token tensor to the beginning of the sequence (and its flipped version) expanded by RMS, allowing Mamba2 to incorporate an overview of the sequence into its hidden state before sequence processing begins. This does not introduce any extra parameters, although it incurs extra FLOPs that equal
%-------------------------------------------------------------------------

\vspace{-1em}
\begin{equation}
    C_{\text{RT}} = \frac{1}{p_t\cdot p_x\cdot p_y}\cdot C_{\text{bimamba}},
\end{equation}
\vspace{-1em}

\noindent
where $p_t,p_y,p_x$ are the average pooling range along the temporal, height, and width dimensions of the video token tensor, respectively. As this equation shows, $C_{\text{RT}}$ also scales linearly in the number of tokens $N$, following the behavior of $C_{\text{bimamba}}$. In practice, we set $\{p_t,p_y,p_x\}=\{8,4,4\}$. Thus, the extra cost of review tokens is marginal.

\subsection{TE-Branch: TEmporal Swin Attention}
\label{sec:tebranch}

\begin{figure}[t]
  \centering
   \includegraphics[width=\linewidth]{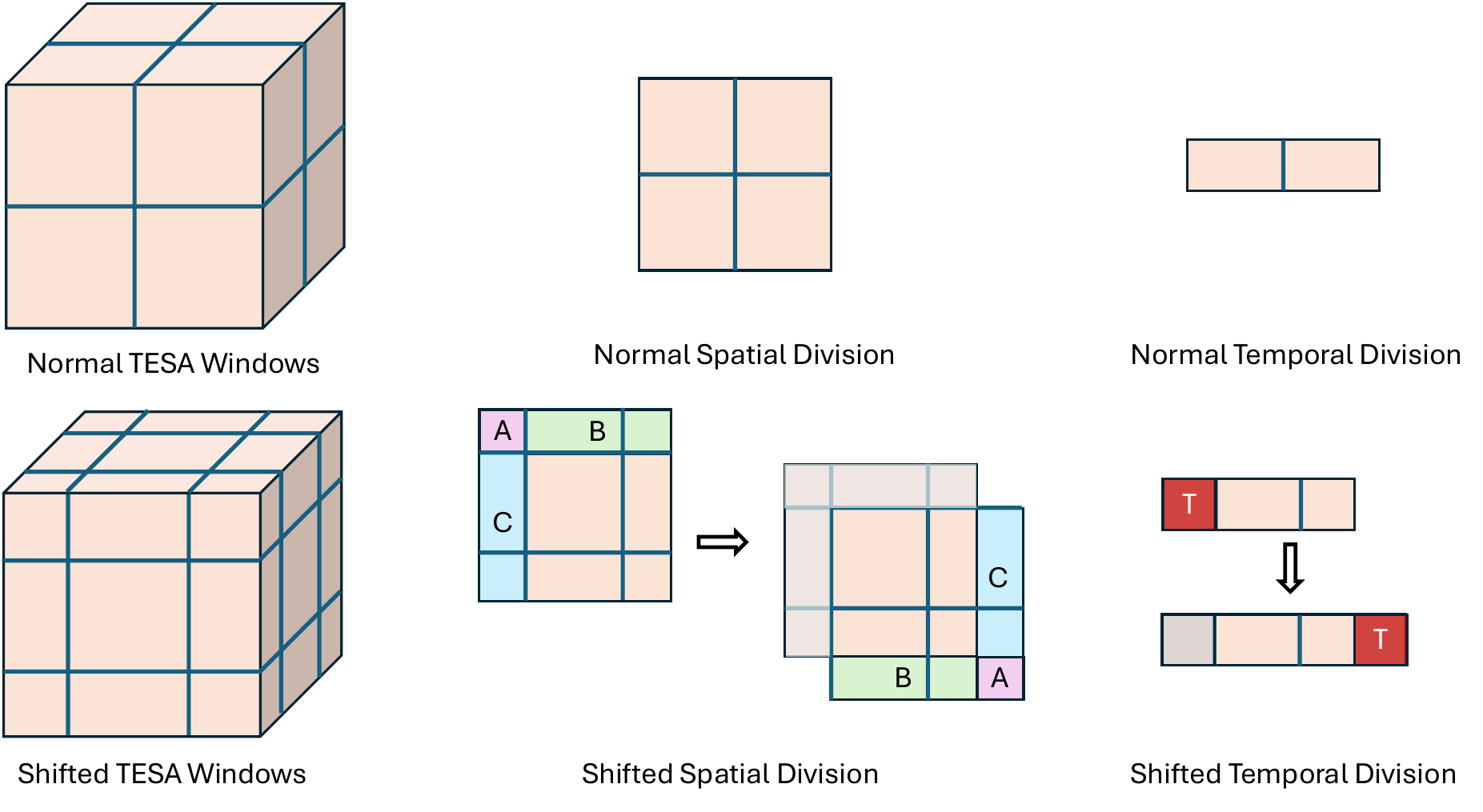}
   \vspace{-2em}
   \caption{\textbf{TEmporal Swin Attention (TESA).} We divide the token tensor into small windows and calculate self-attention within each window. The windows are alternately shifted across layers to cross the boundaries of local windows. The window size remains fixed across different resolutions, hence maintaining linear complexity.}
   \label{fig:tesa}
   \vspace{-1.5em}
\end{figure}

Besides the MA-branch, to further address the adjacency preservation
issue and enhance video consistency, we propose TEmporal Swin Attention (TESA) to build the TE-branch, which gathers short-range information along the spatial dimension and medium-range information along the temporal dimension, as shown in Fig.~\ref{fig:tesa}. It is inspired by a prior window attention work~\cite{liu2021swin}, divides the 3D video token tensor into multiple windows, and calculates attention between tokens within the same window. Assuming the window size is $T_w\times S_w\times S_w$ and the video token tensor size is $T\times H\times W$, the FLOPs of TESA is given by

\vspace{-1.5em}
\begin{equation}
    C_{\text{TESA}}= (8N_w d^2+4N_w^2 d)\cdot \left\lceil \frac{T}{T_w} \right\rceil \cdot \left\lceil \frac{H}{S_w} \right\rceil \cdot \left\lceil \frac{W}{S_w} \right\rceil 
\end{equation}
\vspace{-1em}

\noindent
where $N_w=T_w\cdot S_w\cdot S_w$ and $d$ is the dimension of token embedding vectors. This equation indicates that $C_{\text{TESA}}$ scales linearly in $N=T\cdot H\cdot W$. Its spatial window size $S_w\times S_w$ is very small (we set it to 4$\times$4 in practice), because we mainly use the MA-branch of MATE to deal with spatial correlations and TESA focuses on adjacent correlations along the spatial dimension. Benefiting from such a small spatial window size, TESA incurs negligible extra latency (see Sec.~\ref{sec:ablation}). As indicated in Fig.~\ref{fig:tesa}, the window range of TESA shifts alternatingly in different layers. The self-attention computation in the shifted windows crosses the boundaries of the previous windows, establishing connections among them and enlarging the receptive field. 
% Our experimental results indicate that TESA enhances the consistency of generated videos significantly.

\subsection{Training Recipe}
\label{sec:recipe}

\noindent\textbf{Progressive Training.} We use a progressive recipe (check details in Supp. Mat.) to pre-train our LinGen-4B model. We first pre-train our model on the text-to-image task at a 256p resolution, followed by text-to-video pre-training at progressively higher resolutions (256p to 512p) and longer video lengths (17s to 34s and then 68s). 
% In this progressive training schedule, the token sequence length in the latent space gradually increases. 
% Interestingly, compared to our self-attention baseline, LinGen-4B adapts to longer sequences much more quickly, producing fairly good videos even in the early stages of pre-training (see Sec.~\ref{sec:fastadapt}).

\noindent\textbf{Text-to-Image and Text-to-Video Hybrid Training.} In the text-to-video pre-training stages, we incorporate text-image pairs into the pre-training dataset and perform text-to-image and text-to-video joint training in practice. 
% The sampling ratio of text-image pairs to text-video pairs is 1:100, which is very small, preventing this hybrid setting from reducing the motion of generated videos. 
We find such a hybrid training
% setting not only maintains the model's ability to generate images but also 
improves consistency of generated videos in some failure cases.

\noindent\textbf{Quality Tuning.} Similar to the observation in prior works~\cite{dai2023emu,girdhar2023emuvideo}, we find the quality of generated videos can be greatly enhanced by fine-tuning the model on a small set of high-quality videos. We select 3K high-quality videos from our pre-training dataset and fine-tune our model on them.

%% file: sec/4_experiments.tex
\section{Experiments}
\label{sec:experiments}

In this section, we begin by describing the experimental settings in Sec.~\ref{sec:setting}. We then illustrate the efficiency superiority of LinGen in Sec.~\ref{sec:efficiency}. Next, we benchmark LinGen against state-of-the-art models in Sec.~\ref{sec:sota}. In addition, we demonstrate rapid adaptation of LinGen to longer sequences in Sec.~\ref{sec:fastadapt}.
% highlighting its excellent scalability for generating videos at even higher resolutions and longer lengths. 
Finally, in Sec.~\ref{sec:ablation}, we report on ablation studies that validate the effectiveness of individual modules and techniques incorporated into LinGen. 

\subsection{Experimental Settings}
\label{sec:setting}

\noindent\textbf{Models.} (1) LinGen-4B. We build the denoising module of this model following the setting described in Sec.~\ref{sec:method}. We employ 32 layers with 20 heads in each, with the dimension of embedding vectors being 2560. 
% Flow Matching~\cite{lipman2022flow} and the linear-quadratic t-schedule~\cite{polyak2024moviegen} are used in the denoising process. 
(2) DiT-4B. 
% To validate the superiority of our proposed LinGen, 
We replace MATE blocks in LinGen-4B with global self-attention layers to build a standard DiT. Our DiT-4B has 32 layers with 24 heads in each, with the dimension of embedding vectors being 3072. 
(3) State-of-the-art models. We compare LinGen to state-of-the-art accessible commercial text-to-video generative models, including Runaway Gen3~\cite{runwaygen32024}, Kling~\cite{klingai2024}, and LumaLabs~\cite{lumalabs2024}, and a typical open-source model, OpenSora~\cite{opensora}. We provide comparisons to more open-source models~\cite{yang2024cogvideox,wang2023modelscope,chen2024videocrafter2,wang2023lavie,li2024t2vturbo} in Supp. Mat. Note that most of these open-source models can only generate short videos containing less than 100 raw frames.

\begin{figure}[t]
  \centering
   \includegraphics[width=\linewidth]{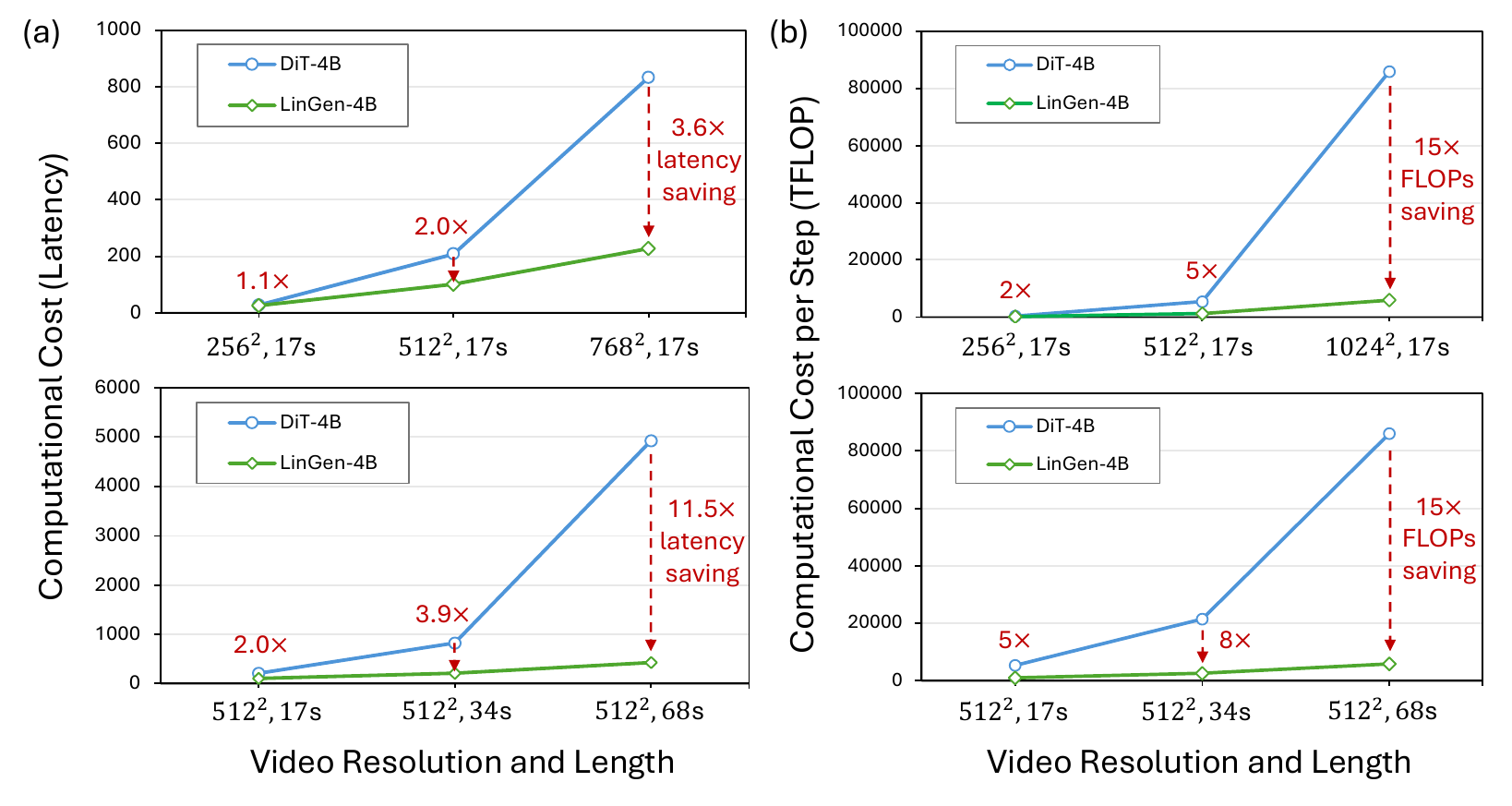}
   \vspace{-2em}
   \caption{\textbf{Computational cost comparison between DiT-4B and LinGen-4B.} (a) Latency. (b) FLOPs. The cost of LinGen scales significantly slower with both video length and video resolution than DiT. Latency is measured on a single H100 GPU.}
   \vspace{-2em}
   \label{fig:saving}
\end{figure}

\noindent\textbf{Datasets.} We use 300M licensed ShutterStock~\cite{shutterstock} text-image pairs and 24M licensed ShutterStock text-video pairs to pre-train our models. We select 3K videos from the ShutterStock and RawFilm~\cite{rawfilm} video dataset to fine-tune our models. More details are provided in Supp. Mat.

\begin{figure*}[t]
  \centering
   \includegraphics[width=\linewidth]{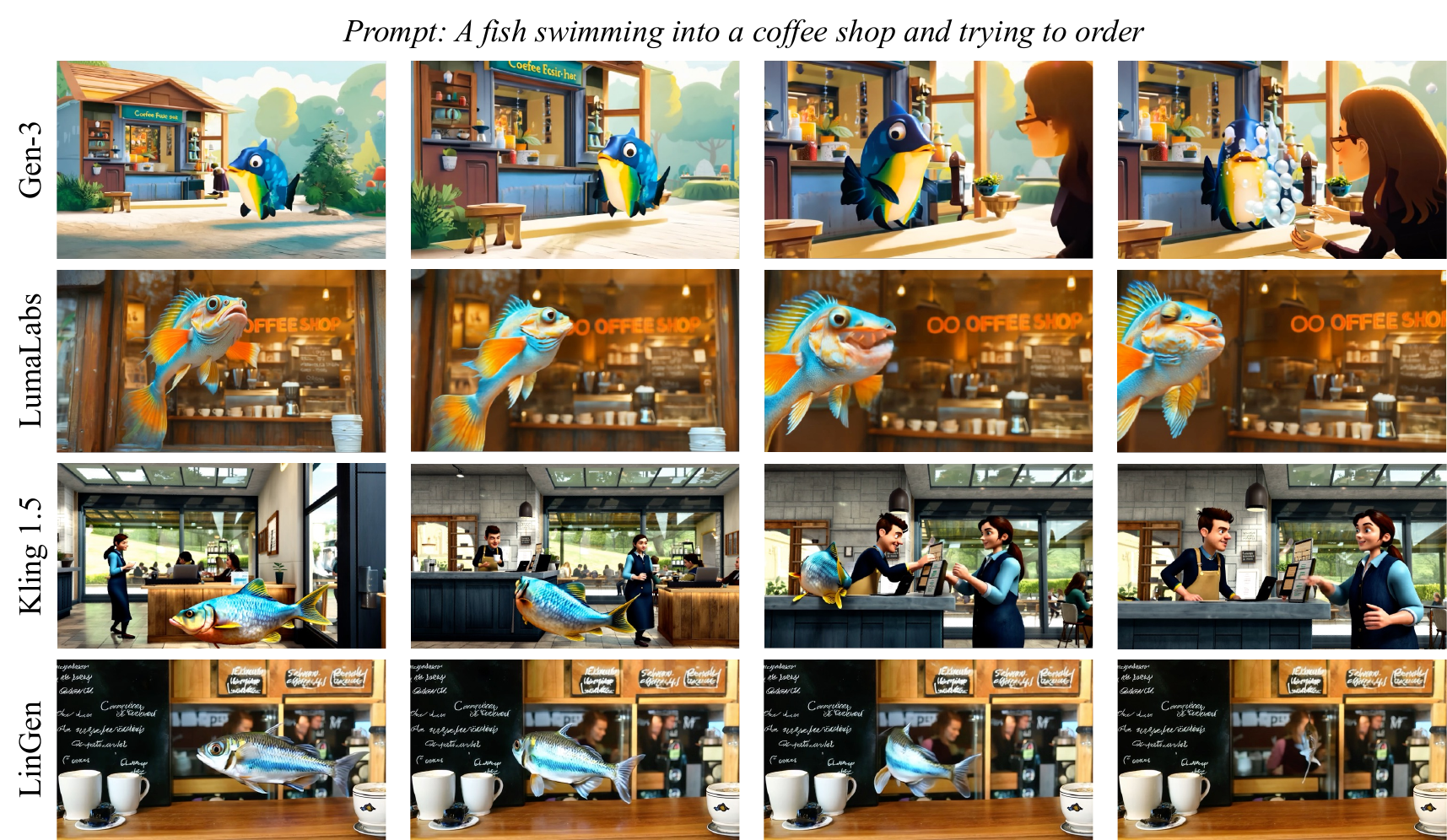}
   \vspace{-1em}
   \caption{Visual examples of videos generated from different models. LinGen-4B generates videos that have similar quality to state-of-the-art commercial video generative models, including Gen-3, LumaLabs, and Kling, while achieving linear complexity and significant speed-up relative to the standard DiT architecture.
   }
   % \vspace{-1em}
   \label{fig:visualexamples}
\end{figure*}

% \subsection{Visual Examples}

% Visual examples are shown in Fig.~\ref{fig:visualexamples}.

\subsection{Efficiency: Linear Computational Complexity}
\label{sec:efficiency}

We compare the efficiency of DiT-4B and our proposed LinGen-4B in terms of FLOPs cost and latency. We show the results in Fig.~\ref{fig:saving}. In terms of FLOPs, LinGen-4B achieves 5$\times$, 8$\times$, and 15$\times$ speed-up relative to DiT-4B when generating 512p videos of 17s, 34s, and 68s length, respectively. In terms of latency, LinGen-4B achieves 2.0$\times$ and 3.6$\times$ speed-up relative to DiT-4B when generating 512p and 768p 17s videos on a single H100, respectively. LinGen-4B achieves 2.0$\times$, 3.9$\times$, and 11.5$\times$ latency speed-up compared to DiT-4B when generating 512p videos of 17s, 34s, and 68s length, respectively. These results indicate that the cost of LinGen scales linearly in the number of pixels in generated videos, thus demonstrating huge efficiency and scalability superiority of LinGen.

\subsection{Comparing Quality to State-of-the-Art Models}
\label{sec:sota}

\begin{table*}
  \centering
  \begin{tabular}{@{}l|>{\centering}p{0.97cm}c>{\centering}p{0.97cm}>{\centering}p{0.97cm}c>{\centering}p{0.97cm}>{\centering}p{0.97cm}>{\centering}p{0.97cm}>{\centering}p{0.97cm}|c@{}}
    \toprule
    \multirow{2}{*}{\textbf{Model}} & Object & Multiple  & Human  & \multirow{2}{*}{Color} & Spatial  & \multirow{2}{*}{Scene} & Appear.  & Temp.  & Overall  & \textbf{Semantic} \\
        & Class & Objects & Action &  & Relatio. & & Style & Style & Consist. & \textbf{Score} \\
    \midrule
    Runway Gen-3~\cite{runwaygen32024} & 87.81\% & 53.64\% & 96.40\% & 80.90\% & 65.09\% & \textbf{54.57\%} & \textbf{24.31\%} & \textbf{24.71\%} & 26.69\% & 75.17\% \\
    Kling~\cite{klingai2024}  & 87.24\% & \textbf{68.05\%} & 93.40\% & 89.90\% & \textbf{73.03\%} & 50.86\% & 19.62\% & 24.17\% & 26.42\% & \textbf{75.68\%} \\
    OpenSora V1.2~\cite{opensora} & 82.22\% & 51.83\% & 91.20\% & \textbf{90.08\%} & 68.56\% & 42.44\% & 23.95\% & 24.54\% & \textbf{26.85\%} & 73.39\% \\
    \midrule
    LinGen & \textbf{90.98\%} & 55.15\% & \textbf{97.50\%} & 83.95\% & 58.15\% & 53.51\% & 21.08\% & 24.29\% & 26.32\% & 73.73\% \\
    \bottomrule
  \end{tabular}
  \begin{tabular}{@{}l|>{\centering}p{0.97cm}>{\centering}p{0.97cm}>{\centering}p{0.97cm}>{\centering}p{0.97cm}>{\centering}p{0.97cm}>{\centering}p{0.97cm}>{\centering}p{0.97cm}|c|c|c@{}}
    \toprule
    \multirow{2}{*}{\textbf{Model}} & Subject  & BG.  & Temp. & Motion  & Aesthe.  & Imag.  & Dyna.  & \textbf{Quality} & \textbf{Total} & \textbf{Max. Raw} \\
    & Consist. & Consis. & Flick. & Smooth. & Quality & Quality & Degree & \textbf{Score} & \textbf{Score} & \textbf{Frames} \\
    \midrule
    Runway Gen-3~\cite{runwaygen32024} & 97.10\% & 96.62\% & 98.61\% & 99.23\% & 60.14\% & \textbf{63.34\%} & \textbf{66.82\%} & \textbf{84.11\%} & \textbf{82.32\%} & 256\\
    Kling~\cite{klingai2024}  & \textbf{98.33\%} & 97.60\% & 99.30\% & \textbf{99.40\%} & 46.94\% & 61.21\% & 65.62\% & 83.39\% & 81.85\% & 313 \\
    OpenSora V1.2~\cite{opensora} & 96.75\% & \textbf{97.61\%} & \textbf{99.53\%} & 98.50\% & 42.39\% & 56.85\% & 63.34\% & 81.35\% & 79.76\% & 408\\
    \midrule
    LinGen & 98.30\% & 97.60\% & 99.26\% & 98.58\% & \textbf{63.67\%} & 60.55\% & 63.36\% & 83.77\% & 81.76\% & \textbf{1088}\\
    \bottomrule
  \end{tabular}
  \vspace{-1em}
  \caption{Automatic evaluation of LinGen on VBench-Long. \textbf{Quality Score} measures the quality of generated videos and \textbf{Semantic Score} measures text-video alignment. \textbf{Total Score} is their weighted sum. Higher values indicate better performance for all these metrics. LinGen is comparable to state-of-the-art commercial models (\ie, Gen-3 and Kling) and outperforms the typical open-source model (\ie, OpenSora) significantly. LinGen not only achieves a much higher maximum number of raw frames but also does so on a single GPU.}
  \vspace{-1.5em}
  \label{tab:vbench}
\end{table*}

We evaluate the performance of our proposed LinGen-4B model and other text-to-video models in three ways: (1) Exhibit visual examples for eyeballing comparison, as shown in Fig.~\ref{fig:visualexamples}. We provide more examples in Supp. Mat. (2) Use human evaluation to perform A/B comparison and calculate win rates. (3) Use automatic quantitative metrics to compare LinGen with more existing text-to-video models. We use a standard video evaluation benchmark, VBench~\cite{huang2023vbench}, to evaluate video quality and text-video faithfulness. VBench comprehensively evaluates text-to-video models using 16 disentangled dimensions. Each dimension is tailored to specific prompts and evaluation methods.

\begin{figure}[t]
  \centering
   \includegraphics[width=\linewidth]{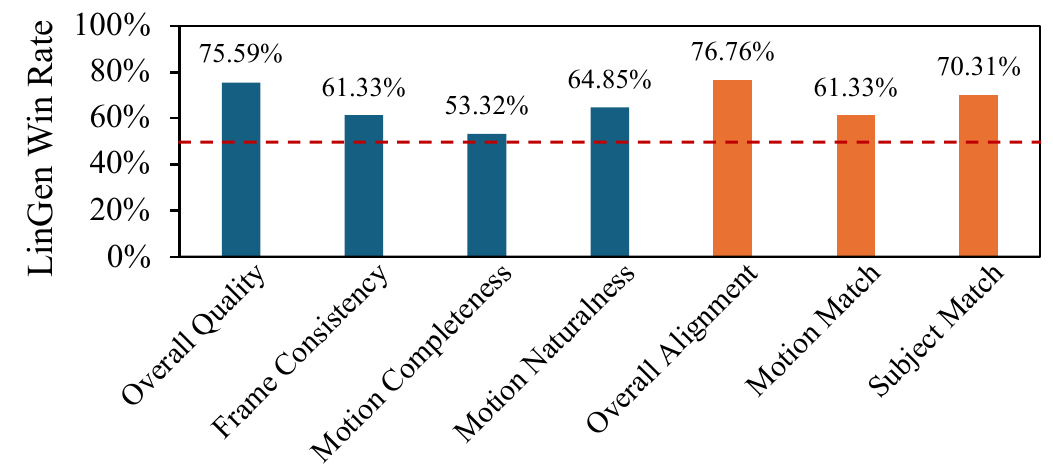}
   \vspace{-2em}
   \caption{Human evaluation on the quality and text-video alignment of videos generated by DiT-4B and LinGen-4B. LinGen outperforms DiT due to it faster adapation to longer token sequences.}
   \vspace{-1em}
   \label{fig:human_attn}
\end{figure}

\begin{figure}[t]
  \centering
   \includegraphics[width=\linewidth]{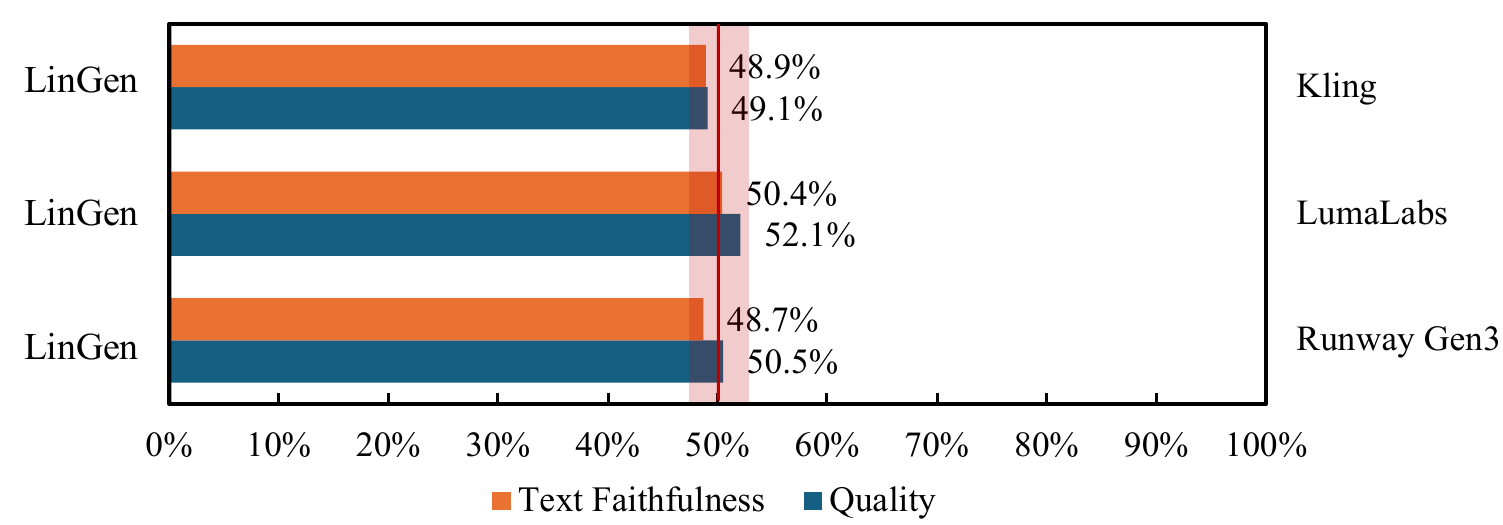}
   \vspace{-2em}
   \caption{Win rates of human evaluation on the quality and text-video alignment of videos generated by LinGen and state-of-the-art video generative models. LinGen has comparable performance to them, given that the variance of human evaluation is 3\%.}
   \vspace{-1.5em}
   \label{fig:human_sota}
\end{figure}

% \begin{figure}[t]
%   \centering
%    \includegraphics[width=\linewidth]{quality_512p.png}
%    \caption{Human evaluation of the quality of 512p videos generated by DiT-4B and LinGen-4B.}
%    \label{fig:quality_512p}
% \end{figure}

% \begin{figure}[t]
%   \centering
%    \includegraphics[width=\linewidth]{human_eval_other.png}
%    \caption{Win rates of human evaluation in the comparison between LinGen and prior text-to-video models. Videos are generated at 512p resolution.}
%    \label{fig:human_eval_other}
% \end{figure}

\noindent\textbf{Human Evaluation Results.} We compare the quality and text-faithfulness of videos generated by DiT-4B and LinGen-4B at 256p after being trained for 40K steps with a batch size of 1024; results are shown in Fig.~\ref{fig:human_attn}. This indicates that \textbf{LinGen-4B outperforms DiT-4B in both video quality and text-video alignment}, while achieving linear complexity and significant speed-up. We speculate that, while both models are transferred from the text-to-image generation task, LinGen exhibits a superior ability to adapt to longer token sequences (see Sec.~\ref{sec:fastadapt}). Consequently, LinGen learns text-to-video generation more efficiently than DiT, resulting in improved performance. Fig.~\ref{fig:human_sota} incidates that LinGen has comparable performance to state-of-the-art commerical video generative models.
% The variance of human evaluation on overall quality, frame consistency, motion completeness, and motion naturalness is 6\%, 5.9\%, 3.2\%, and 0.8\%, respectively. 
% This means that the generated video quality of LinGen-4B is comparable to that of DiT-4B in almost all aspects at 256p and 512p resolutions. 
% We also compare LinGen to representative accessible text-to-video models via human evaluation, including T2V-Turbo, VideoCrafter2, Gen2, and Imagen Video, as shown in Fig.~\ref{fig:human_eval_other}. Note that they cannot generate videos as long as 17s; hence, we use their generated videos of maximum length to compare with LinGen.

\noindent\textbf{Automatic Quantitative Results.} 
% We evaluate LinGen on VBench with the MovieGen Bench prompt sets~\cite{polyak2024moviegen}. 
Given that the shortest video from LinGen is 17s long, significantly surpassing most models on the VBench-standard leaderboard, we evaluate LinGen against models on VBench-Long instead, as shown in Table~\ref{tab:vbench}. 
It shows that \textbf{LinGen outperforms Kling in terms of video quality and has similar overall performance to both Gen-3 and Kling, while achieving linear complexity and enabling more than one thousand raw frames generation on a single GPU.} LinGen outperforms OpenSora significantly. 
We provide the complete leaderboard and evaluation results on VBench-standard and VBench-Custom in Supp. Mat.
% Note that the aesthetic quality of LinGen is lower than expected because VBench uses the LAION aesthetic model~\cite{schuhmann2022laion} to assign aesthetic scores to generated videos, and this model prefers the cartoon style to the photorealistic style. In terms of imaging quality, LinGen outperforms all these models significantly.

\subsection{Adaptation to Longer Token Sequences}
\label{sec:fastadapt}

\begin{figure}[t]
  \centering
   \includegraphics[width=\linewidth]{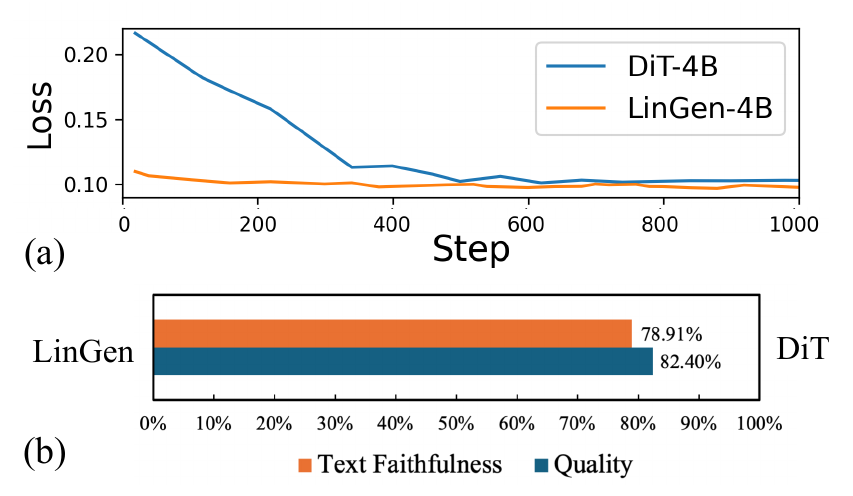}
   \vspace{-2.5em}
   \caption{LinGen adapts much faster to the new task than DiT. (a) Loss curves when transferring the model trained on 256p video generation to 512p. (b) Win rates of human evaluation on quality and text-video faithfulness comparison between LinGen-4B and DiT-4B. Checkpoints are selected after 1K pre-training steps.}
   \vspace{-1em}
   \label{fig:adaptation}
\end{figure}

% \begin{figure}[t]
%   \centering
%    \includegraphics[width=\linewidth]{adapt_human.png}
%    \caption{Win rates of human evaluation on quality comparison between LinGen-4B and DiT-4B. Checkpoints are selected after 1K pre-training steps.}
%    \label{fig:adapt_human}
% \end{figure}

LinGen adapts to longer sequences of latent tokens more quickly than DiT. This could benefit from the strong adaptation ability of Mamba models to longer sequences, which has also been observed in language tasks~\cite{ren2024samba}. We observe this phenomenon in the loss curves when transferring the model trained on 256p video generation to 512p generation in progressive training, as shown in Fig.~\ref{fig:adaptation} (a). We further conduct a human evaluation on the checkpoints at an early stage of 512p 17s video generation pre-training and 512p 34s video generation pre-training, as shown in Fig.~\ref{fig:adaptation} (b). The results validate our observation that LinGen adapts more quickly to longer sequences of latent tokens than DiT, which means better scalability for video generation at higher resolutions and longer lengths.

\subsection{Ablation Experiments}
\label{sec:ablation}

\begin{figure}[t]
  \centering
   \includegraphics[width=\linewidth]{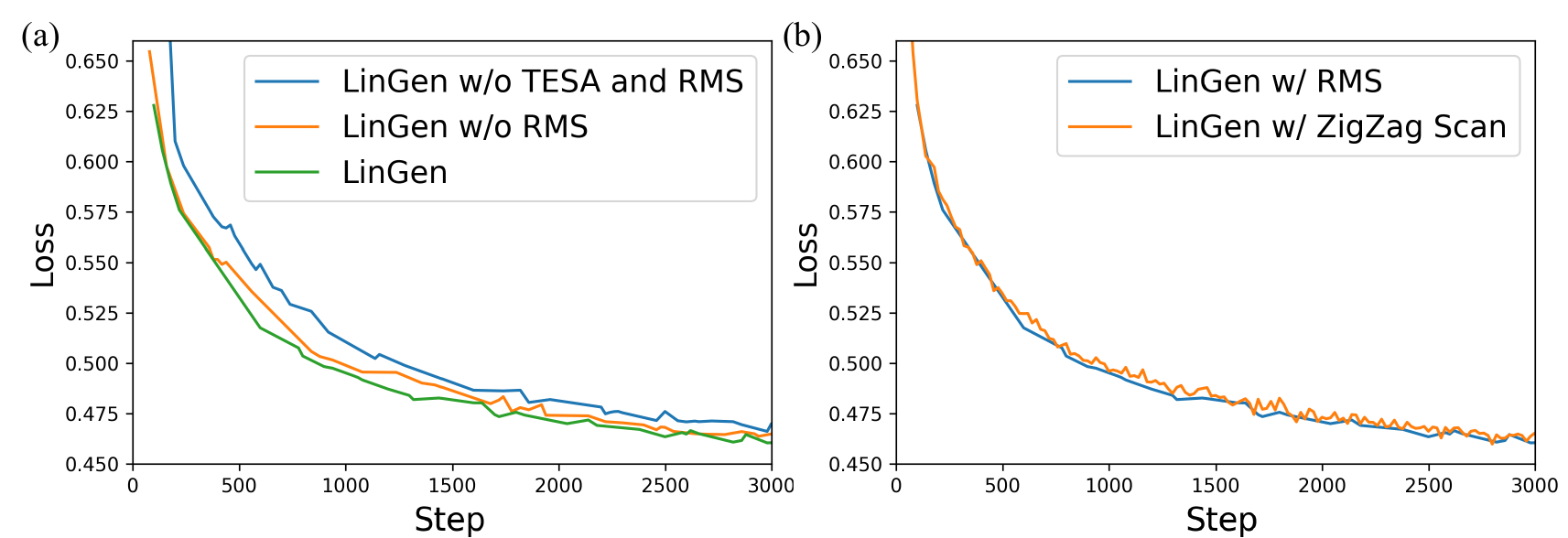}
   \vspace{-2em}
   \caption{Loss curves of 256p text-to-video pre-training under different settings. (a) Ablation on the TESA block and RMS. (b) Ablation on different scan methods.}
   \vspace{-2em}
   \label{fig:ablation_loss}
\end{figure}

\begin{figure}[t]
  \centering
   \includegraphics[width=\linewidth]{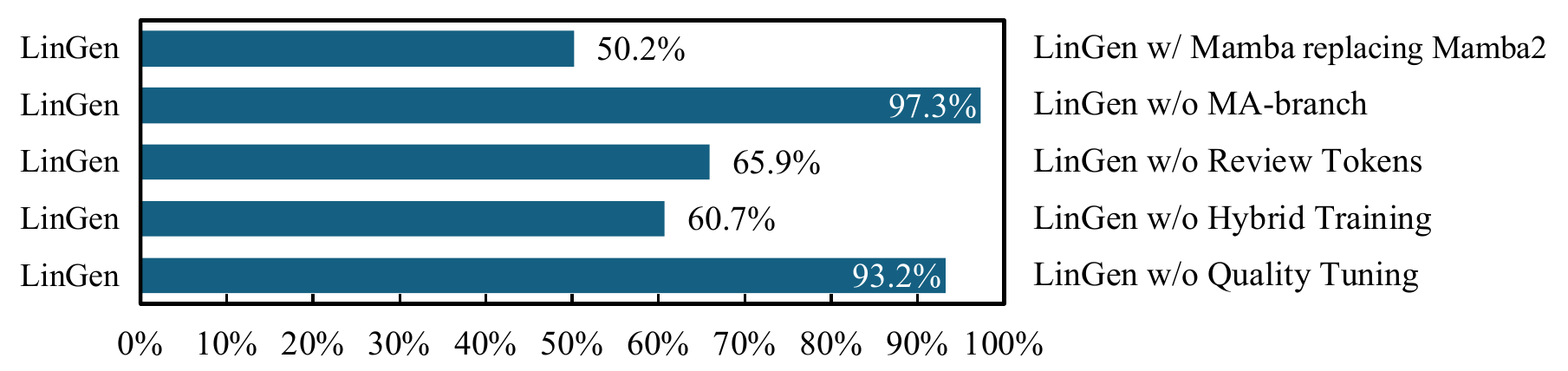}
   \vspace{-2em}
   \caption{Win rates of human evaluation on quality comparison between the LinGen default setting and corresponding variants.}
   \vspace{-1em}
   \label{fig:ablation}
\end{figure}

\textbf{For performance}, we conduct ablation experiments on the 256p 17s video generation task in two ways: (1) Comparing loss curves. The prior work~\cite{polyak2024moviegen} has observed that the loss curve correlates well with visual quality evaluated by humans. Thus, we compare the loss curves under different training settings to validate their effectiveness, as shown in Fig.~\ref{fig:ablation_loss}. (2) Performing human evaluations. We select corresponding checkpoints after 30K pre-training steps and perform A/B quality comparison between the default setting of LinGen and the changed setting of LinGen. The win rates are shown in Fig.~\ref{fig:ablation}. We provide more visual examples in Supp. Mat. \textbf{For efficiency}, we measure 512p 17s video generation latency of LinGen under different settings, as shown in Table~\ref{tab:ablation_latency}. Fig.~\ref{fig:ablation} validates the effectiveness of review tokens, hybrid training, and quality tuning, and Table~\ref{tab:ablation_latency} shows review tokens incur marginal extra latency.

\begin{table}
  \centering
  \scriptsize
  \begin{tabular}{c|ccc|c}
    \toprule
    \multirow{2}{*}{\textbf{TE-branch}} & \multicolumn{3}{c|}{\textbf{MA-branch}} & \multirow{2}{*}{\textbf{Latency/s}}  \\
        & Scan Method & Mamba version & Review Tokens &  \\
    \midrule
    \rowcolor{blue!20}
    \ding{51} & RMS & Mamba2 & \ding{51}  & 102 \\
    \cellcolor{gray!30}\ding{55} & RMS & Mamba2 & \ding{51} & 94 (-8) \\
    \ding{51} & \cellcolor{gray!30}\ding{55} & Mamba2 & \ding{51}  & 99 (-3) \\
    \ding{51} & \cellcolor{gray!30}Zigzag & Mamba2 & \ding{51} & 144 (+42)\\
    \ding{51} & RMS & \cellcolor{gray!30}Mamba & \ding{51}  & 127 (+25)\\
    \ding{51} & RMS & Mamba2 & \cellcolor{gray!30}\ding{55}   & 98 (-4) \\
    \ding{51} & \cellcolor{gray!30}\ding{55} & \cellcolor{gray!30}\ding{55} & \cellcolor{gray!30}\ding{55} & 65 (-37)\\
    \bottomrule
  \end{tabular}
  \vspace{-0.5em}
  \caption{Latency of the LinGen default setting and variant settings when generating 512p 17s videos.}
  \vspace{-2em}
  \label{tab:ablation_latency}
\end{table}

\noindent\textbf{TESA Block.} Table~\ref{tab:ablation_latency} shows that the TESA block only incurs marginal latency, while Fig.~\ref{fig:ablation_loss} indicates that it contributes effectively to the quality of generated videos. As expected, TESA is efficient due to its small window size, while being very effective due to its addressing of the adjacency preservation issue and enhancing medium-range temporal correlation calculation.

\noindent\textbf{Rotary Major Scan.} Fig.~\ref{fig:ablation_loss} (a) shows that RMS is effective in improving video quality by mitigating the adjacency preservation issue, while causing negligible extra latency, as indicated by Table~\ref{tab:ablation_latency}. On the contrary, existing scan methods, such as Zigzag scan, incur a significant latency increment when operating on huge 3D video token tensors. In addition, we find the loss curve of LinGen w/ Zigzag scan is almost the same as that of LinGen w/ RMS, as shown in Fig.~\ref{fig:ablation_loss} (b), indicating RMS achieves similar performance to Zigzag scan with a much lower extra latency.

\noindent\textbf{Mamba and Mamba2.} Compared to Mamba, Mamba2 is more efficient and hardware-friendly~\cite{dao2024mamba2}. Table~\ref{tab:ablation_latency} validates this, showing that LinGen w/ Mamba2 is 25\% faster than LinGen w/ Mamba. 
% Although Mamba2 compromises expressive power due to simplification of the decay matrix in an SSM~\cite{dao2024mamba2}, it compensates for this using a much larger hidden state size. We set the hidden state size to 16 and 128 in LinGen w/ Mamba and LinGen w/ Mamba2, respectively, for both quality comparison and latency measurement, following their default values in the original design~\cite{dao2024mamba2}. 
Fig.~\ref{fig:ablation} shows that LinGen w/ Mamba2 achieves almost the same video quality as LinGen w/ Mamba. In addition, although giving up the whole MA-branch brings significant speed-up, it severely impacts the quality of generated videos, as shown in Table~\ref{tab:ablation_latency} and Fig.~\ref{fig:ablation}, proving the necessity of including the MA-branch.

% \noindent\textbf{Review Tokens.} Table~\ref{tab:ablation_latency} shows that review tokens only incur marginal extra latency, while human evaluation results in Fig.~\ref{fig:ablation} validates the effectiveness of review tokens on improving the quality of generated videos.

% \noindent\textbf{Hybrid Training and Quality Tuning.} 
% We incorporate text-image pairs into the training dataset and perform text-to-image and text-to-video hybrid training in text-to-video pre-training stages. We find this not only maintains the image generation ability of LinGen but also avoids failure cases in which the generated video is very inconsistent. 
% Human evaluation results in Fig.~\ref{fig:ablation} show that both (1) text-to-image and text-to-video hybrid training and (2) fine-tuning our model on a small high-quality dataset contributes effectively to the quality improvement of generated videos. 

% \noindent\textbf{Quality Tuning.} We fine-tune our model on a small set of videos with extremely high quality and good motions, after pre-training. We observe that this improves the fidelity and motion of generated videos significantly, which is validated by the human evaluation results shown in Fig.~\ref{fig:ablation}.

%% file: sec/5_conclusion.tex
\section{Conclusion}
\label{sec:conclusion}

In the paper, we proposed LinGen, a linear-complexity text-to-video generation framework that enables high-resolution minute-length video generation on a single GPU. It replaces self-attention layers in DiTs with our novel MATE block, which inherits linear complexity from its two branches: MA-branch and TE-branch. Compared to the native Mamba block, MATE addresses its adjacency preservation issue and comprehensively enhances short-, medium-, and long-range correlations, improving the consistency and fidelity of generated videos significantly. Our experimental results show that LinGen achieves linear complexity and up to 11.5$\times$ speed-up in terms of latency, while maintaining the high quality of generated videos. LinGen presents a linear-complexity self-attention replacement, 
% that can effectively handle the complex, dense prediction task of text-to-video generation, 
paving the way for broader adoption of this framework to hour-length video generation and real-time interactive video generation.
% linear alternatives to quadratic-complexity self-attention across various fields. 

% Looking ahead, we see several valuable directions to explore: (1) Further enhancing video generation efficiency through sampling distillation to enable real-time interactive video generation. Infinite-length video generation can potentially be achieved with a denoising pipeline~\cite{kodaira2023streamdiffusion}. (2) Further extending video length to generate hour-long videos. Creating such videos would require handling text prompts as lengthy as articles, placing high demands on the capacity of text encoders to maintain effective content control over the generated videos.

\section*{Acknowledgment}

This work was supported in part by a Meta summer internship and in part by NSF under Grant No. CCF-2203399.

%% file: sec/X_suppl.tex
\clearpage
\setcounter{page}{1}
\maketitlesupplementary
\appendix

\section{Adjacency Preservation}
\label{app:adjacency}

\begin{figure}[h]
  \centering
  % \vspace{-1em}
  \includegraphics[width=\linewidth]{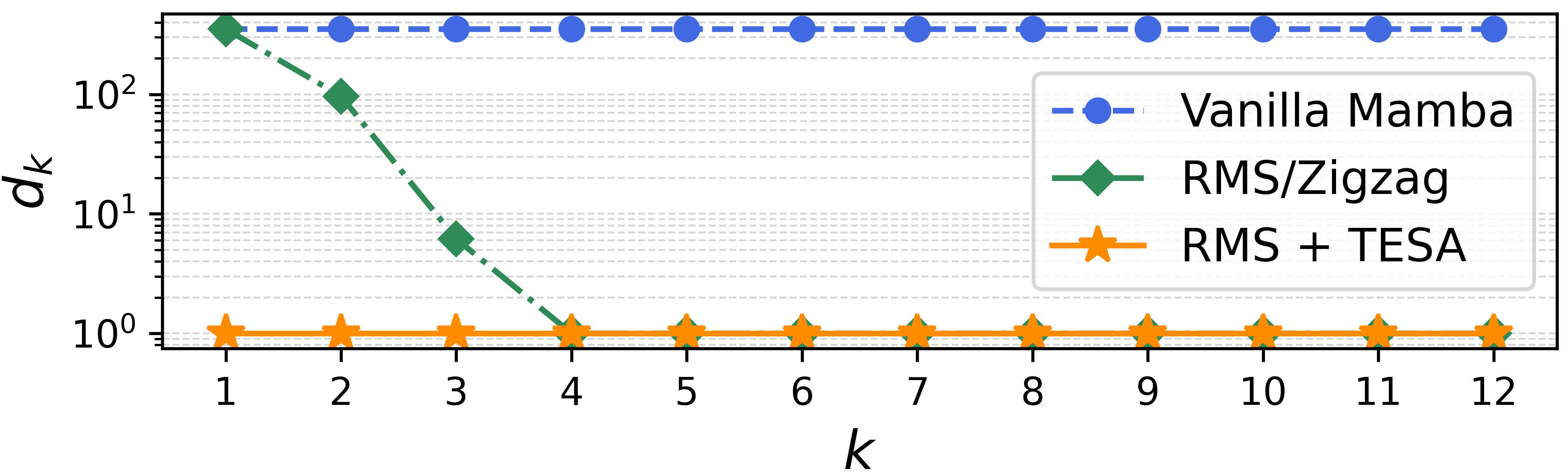}
  % \vspace{-2.5em}
   \caption{Average minimum distance between adjacent tokens.}
   % \vspace{-1em}
   \label{fig:distance}
\end{figure}

Vanilla Mamba2 cannot be scaled to process huge images and video tokens well due to its long-range decay and the well-known adjacency preservation issue (see Sec.~1 of the main paper), causing distorted and inconsistent videos. Loss comparisons in Fig.~11 of the main paper and ablative videos (see Sec.~\ref{sec:app_examples}) validate the effectiveness of RMS and TESA. 
Mamba models scan image and video tokens into a sequence, where the \textbf{\textit{minimum distance}} between originally adjacent tokens in $k$ layers reflects their most precise correlation. For an $H\times W\times T$ token tensor, 
we compute \textbf{\textit{its average}}, $d_k$, among adjacent tokens in a $2\times 2\times 2$ cube and plot $d_k$ in Fig.~\ref{fig:distance} for $H=W=T=32$. RMS achieves the same $d_k$ as Zigzag while being much more efficient and scalable (see Table 2 of the main paper). RMS and TESA thoroughly address the adjacency preservation issue.

\section{Visual Examples}
\label{sec:app_examples}

We provide visual examples next. They can also be found on our \href{https://lineargen.github.io/}{project website}.

\begin{figure*}[t]
  \centering
   \includegraphics[width=\linewidth]{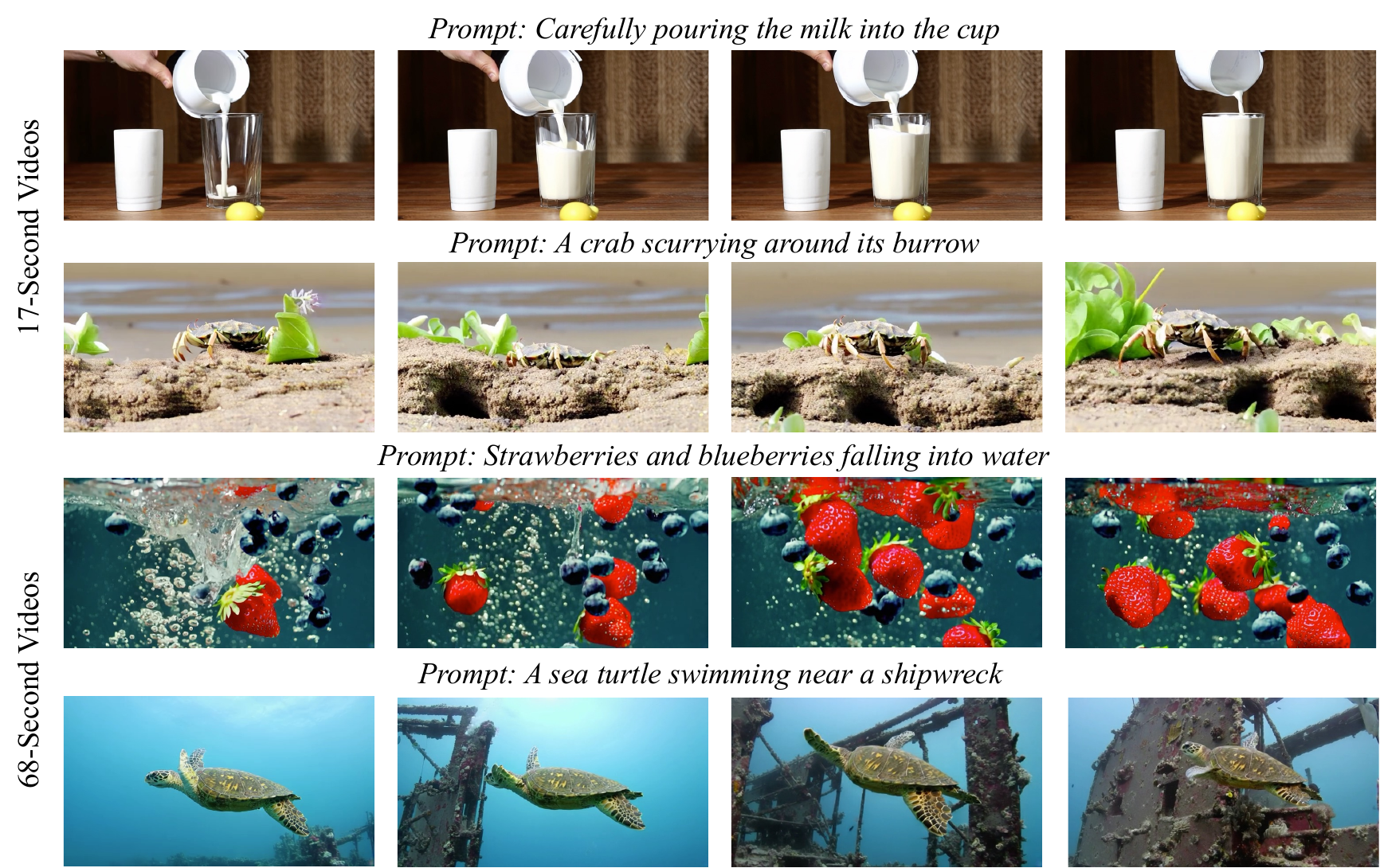}
   \caption{Examples of 17-second and 68-second videos generated by LinGen.}
   \label{fig:demovideo}
\end{figure*}

\begin{figure*}[t]
  \centering
   \includegraphics[width=\linewidth]{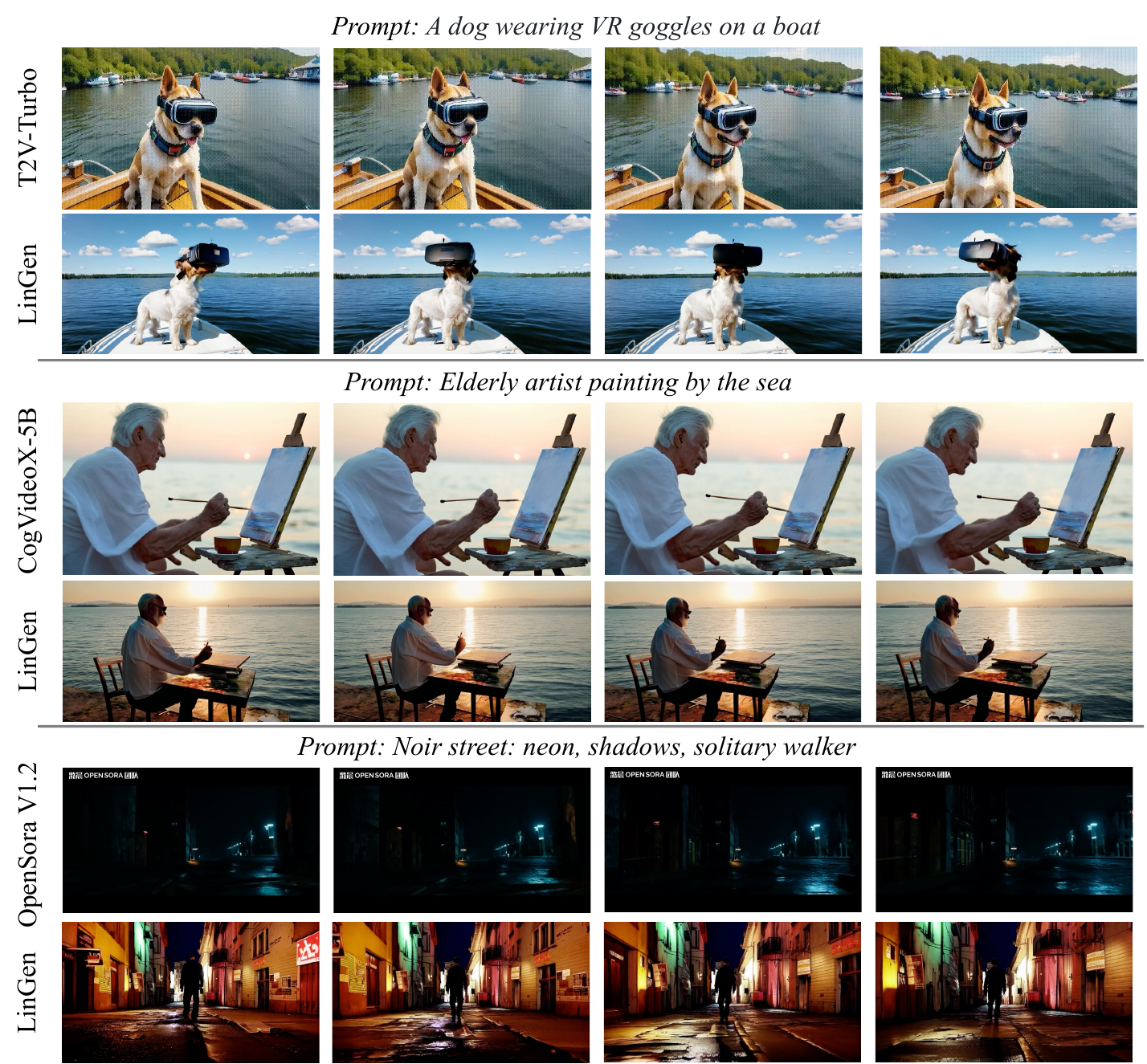}
   \caption{Comparisons with typical open-source video generative models.}
   \label{fig:opcomp}
\end{figure*}

\begin{figure*}[t]
  \centering
   \includegraphics[width=\linewidth]{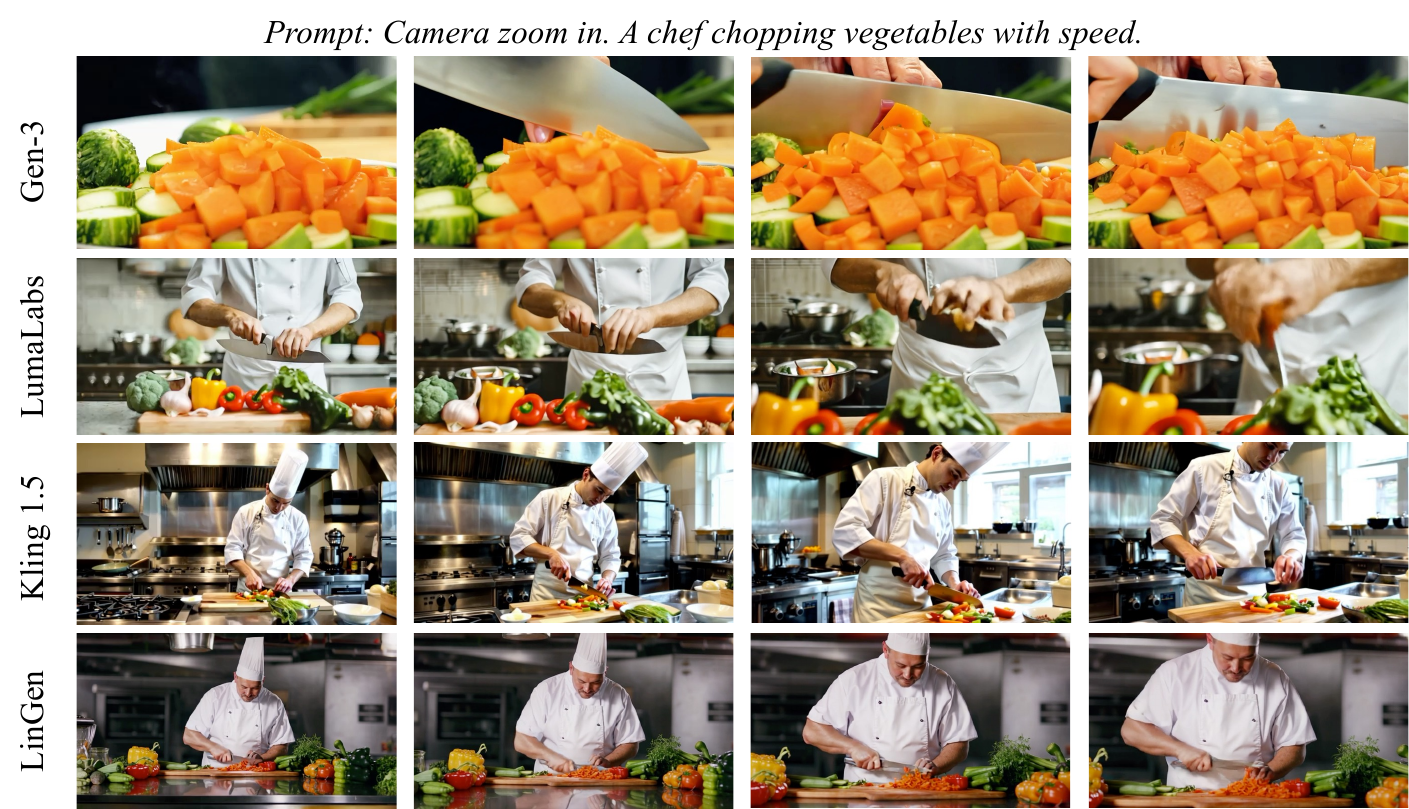}
   \caption{Comparisons with state-of-the-art accessible commercial models.}
   \label{fig:sotacomp}
\end{figure*}

\begin{figure*}[t]
  \centering
   \includegraphics[width=\linewidth]{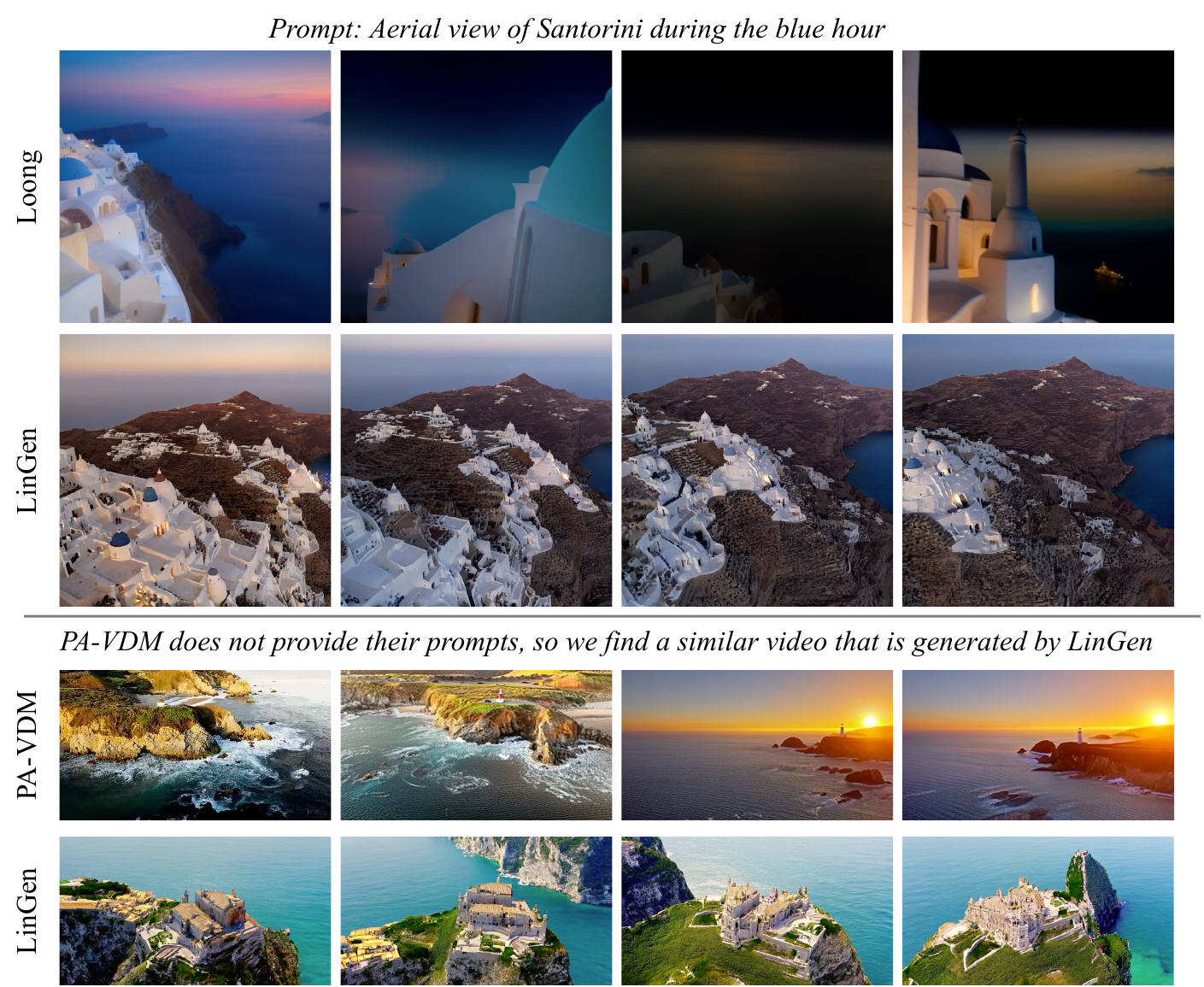}
   \caption{Comparisons with existing trials on generating minute-length videos.}
   \label{fig:minutecomp}
\end{figure*}

\begin{figure*}[t]
  \centering
   \includegraphics[width=\linewidth]{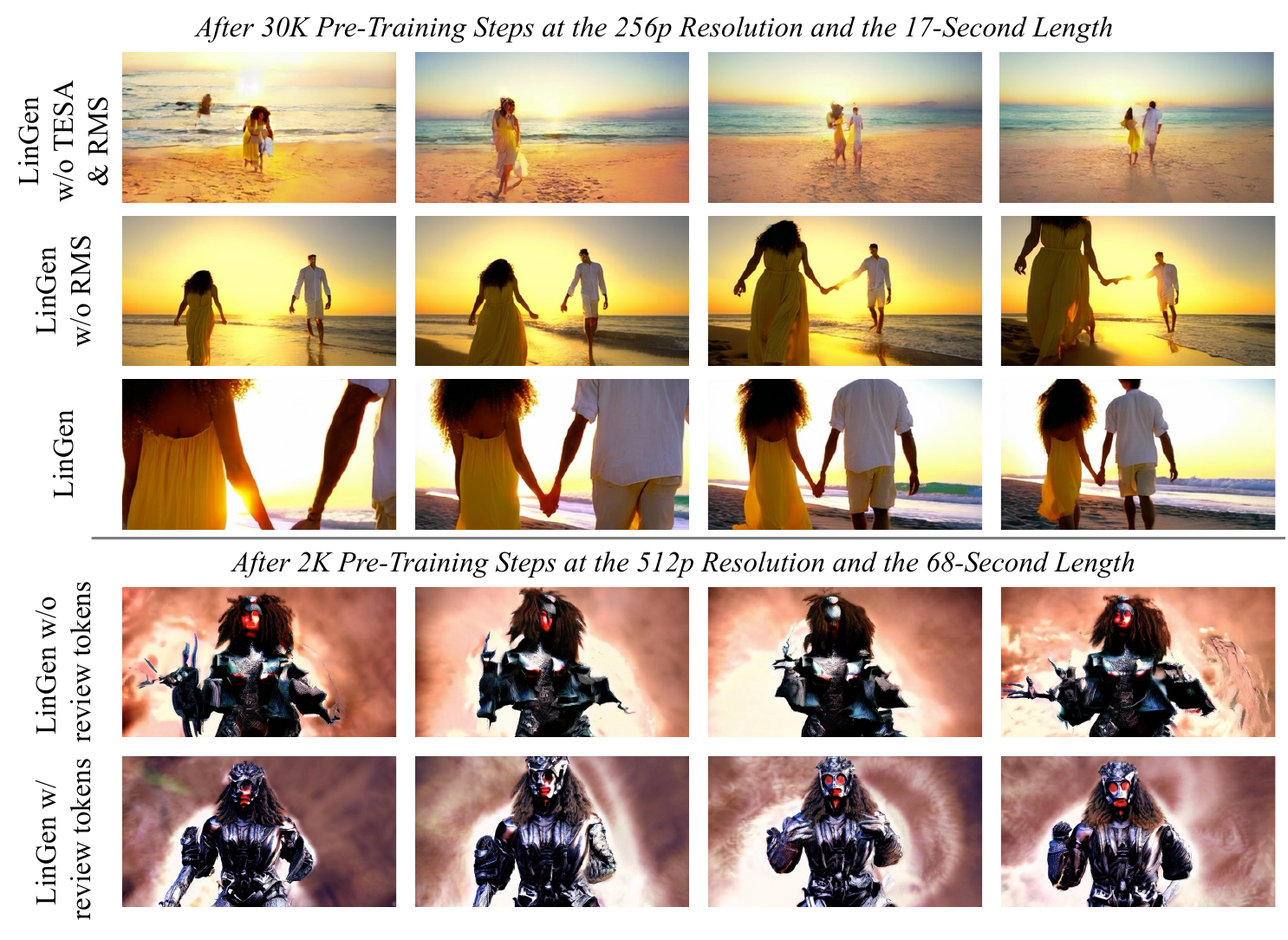}
   \caption{Visual examples of ablation experiments on the TESA block, RMS, and review tokens.}
   \label{fig:ablation_module}
\end{figure*}

\begin{figure*}[t]
  \centering
   \includegraphics[width=\linewidth]{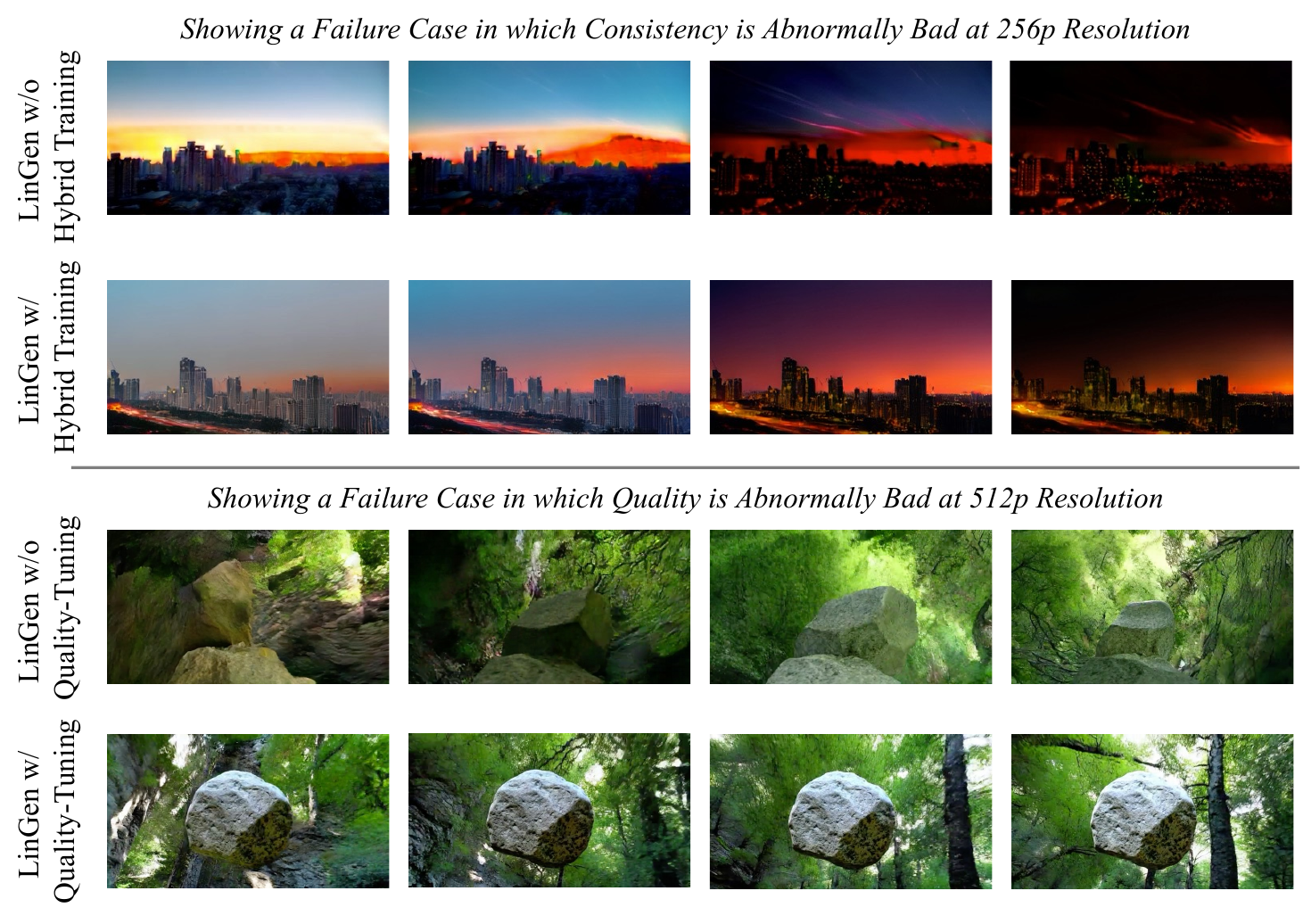}
   \caption{Visual examples of ablation experiments on hybrid training and quality-tuning.}
   \label{fig:ablation_tech}
\end{figure*}

\begin{itemize}
    \item \textbf{Video Demos.} 17-second and 68-second videos generated by LinGen (see Fig.~\ref{fig:demovideo}).
    \item \textbf{Comparisons with existing video generation works.} Our baselines are typical open-source models (see Fig.~\ref{fig:opcomp}), including T2V-Turbo-v2~\cite{li2024t2vturbo}, CogVideoX-5B~\cite{yang2024cogvideox}, and OpenSora v1.2~\cite{opensora}, state-of-the-art accessible commercial models (see Fig.~\ref{fig:sotacomp}), including Kling~\cite{klingai2024}, Runway Gen3~\cite{runwaygen32024}, and LumaLabs~\cite{lumalabs2024}, and minute-length video generation trials (see Fig.~\ref{fig:minutecomp}), including Loong~\cite{wang2024loong} and PA-VDM~\cite{xie2024progressive}. Note that PA-VDM has not yet released the code and prompts. Thus, we selected one LinGen-generated video similar to their demo video for reference.
    \item \textbf{Ablation experiments.} Video comparisons to validate the effectiveness of modules and techniques deployed in LinGen, including TEmporal Swin Attention (TESA), Rotary-Major Scan (RMS), review tokens, hybrid training, and quality-tuning (see Fig.~\ref{fig:ablation_module} and Fig.~\ref{fig:ablation_tech}).
\end{itemize}

\section{Comparisons with Prior Works}

In this section, we first supplement VBench results reported in Sec.~\ref{sec:app_vbench} in order to compare with more models and discuss the limitations of VBench. Then, we present visual examples of the generated videos to provide comparisons with prior works and include additional human evaluation results in Sec.~\ref{sec:app_viz} to demonstrate high quality of videos generated by LinGen.

\subsection{Automatic Metrics: VBench Results}
\label{sec:app_vbench}

\begin{table*}
  \centering
  \begin{tabular}{@{}l|>{\centering}p{0.97cm}>{\centering}p{0.97cm}>{\centering}p{0.97cm}>{\centering}p{0.97cm}>{\centering}p{0.97cm}>{\centering}p{0.97cm}>{\centering}p{0.97cm}|c|c|c@{}}
    \toprule
    \multirow{2}{*}{\textbf{Model}} & Subject  & BG.  & Temp. & Motion  & Aesthe.  & Imag.  & Dyna.  & \textbf{Quality} & \textbf{Total} & \textbf{Max. Raw} \\
    & Consist. & Consis. & Flick. & Smooth. & Quality & Quality & Degree & \textbf{Score} & \textbf{Score} & \textbf{Frames} \\
    \midrule
    Runway Gen-3~\cite{runwaygen32024} & 97.10\% & 96.62\% & 98.61\% & 99.23\% & 60.14\% & \textbf{63.34\%} & \textbf{66.82\%} & \textbf{84.11\%} & \textbf{82.32\%} & 256\\
    Kling~\cite{klingai2024}  & \textbf{98.33\%} & 97.60\% & 99.30\% & \textbf{99.40\%} & 46.94\% & 61.21\% & 65.62\% & 83.39\% & 81.85\% & 313 \\
    CogVideoX-5B~\cite{yang2024cogvideox} & 96.23\%  & 96.52\%	& 98.66\%	& 96.92\%	& \textbf{70.97\%}	& 61.98\%	& 62.90\%	& 82.75\%	& 81.61\% & 48 \\
    Mochi-1~\cite{genmo2024mochi} & 96.99\%	& 97.28\%	& 99.40\%	& 99.02\%	& 61.85\%	& 56.94\% &	60.64\% &	82.64\% &	80.13\% & 163 \\
    OpenSora V1.2~\cite{opensora} & 96.75\% & \textbf{97.61\%} & \textbf{99.53\%} & 98.50\% & 42.39\% & 56.85\% & 63.34\% & 81.35\% & 79.76\% & 408\\
    Mira~\cite{ju2024miradata} & 96.23\%	& 96.92\%	& 98.29\%	& 97.54\%	& 60.33\%	& 42.51\%	& 60.16\%	& 78.78\%	& 71.87\% & 60 \\
    \midrule
    LinGen & 98.30\% & 97.60\% & 99.26\% & 98.58\% & 63.67\% & 60.55\% & 63.36\% & 83.77\% & 81.76\% & \textbf{1088}\\
    \bottomrule
  \end{tabular}
  \begin{tabular}{@{}l|>{\centering}p{0.97cm}c>{\centering}p{0.97cm}>{\centering}p{0.97cm}c>{\centering}p{0.97cm}>{\centering}p{0.97cm}>{\centering}p{0.97cm}>{\centering}p{0.97cm}|c@{}}
    \toprule
    \multirow{2}{*}{\textbf{Model}} & Object & Multiple  & Human  & \multirow{2}{*}{Color} & Spatial  & \multirow{2}{*}{Scene} & Appear.  & Temp.  & Overall  & \textbf{Semantic} \\
        & Class & Objects & Action &  & Relatio. & & Style & Style & Consist. & \textbf{Score} \\
    \midrule
    Runway Gen-3~\cite{runwaygen32024} & 87.81\% & 53.64\% & 96.40\% & 80.90\% & 65.09\% & \textbf{54.57\%} & 24.31\% & \textbf{24.71\%} & 26.69\% & 75.17\% \\
    Kling~\cite{klingai2024}  & 87.24\% & \textbf{68.05\%} & 93.40\% & 89.90\% & \textbf{73.03\%} & 50.86\% & 19.62\% & 24.17\% & 26.42\% & 75.68\% \\
    CogVideoX-5B~\cite{yang2024cogvideox} & 85.23\%	& 62.11\%	& \textbf{99.40\%}	& 82.81\%	& 66.35\%	& 53.20\%	& \textbf{24.91\%}	& 25.38\%	& \textbf{27.59\%}	& \textbf{77.04\%} \\
    Mochi-1~\cite{genmo2024mochi} & 86.51\%	& 50.47\%	& 94.60\%	& 79.73\%	& 69.24\%	& 36.99\%	& 20.33\%	& 23.65\%	& 25.15\% &	70.08\%\\
    OpenSora V1.2~\cite{opensora} & 82.22\% & 51.83\% & 91.20\% & \textbf{90.08\%} & 68.56\% & 42.44\% & 23.95\% & 24.54\% & 26.85\% & 73.39\% \\
    Mira~\cite{ju2024miradata} & 52.06\%	& 12.52\%	& 63.80\%	&42.24\%	&27.83\%	&16.34\%	& 21.89\%	&18.77\%	&18.72\%	&44.21\%\\
    \midrule
    LinGen & \textbf{90.98\%} & 55.15\% & 97.50\% & 83.95\% & 58.15\% & 53.51\% & 21.08\% & 24.29\% & 26.32\% & 73.73\% \\
    \bottomrule
  \end{tabular}
  \caption{A more complete \textbf{VBench-Long} leaderboard. \textbf{Quality Score} measures the quality of generated videos and \textbf{Semantic Score} measures text-video alignment. \textbf{Total Score} represents their weighted sum. Higher values indicate better performance for all these metrics. LinGen can be seen to be comparable to state-of-the-art commercial models (\ie, Gen-3 and Kling) and significantly outperform typical open-source models.}
  \label{tab:vbench_long_comp}
\end{table*}

\begin{table*}
  \centering
  \begin{tabular}{@{}l|>{\centering}p{0.97cm}>{\centering}p{0.97cm}>{\centering}p{0.97cm}>{\centering}p{0.97cm}>{\centering}p{0.97cm}>{\centering}p{0.97cm}>{\centering}p{0.97cm}|c|c|c@{}}
    \toprule
    \multirow{2}{*}{\textbf{Model}} & Subject  & BG.  & Temp. & Motion  & Aesthe.  & Imag.  & Dyna.  & \textbf{Quality} & \textbf{Total} & \textbf{Max. Raw} \\
    & Consist. & Consis. & Flick. & Smooth. & Quality & Quality & Degree & \textbf{Score} & \textbf{Score} & \textbf{Frames} \\
    \midrule
    T2V-Turbo-v2~\cite{li2024t2vturbov2}      & 95.50\% & 96.71\% & 97.35\% & 97.07\% & \textbf{90.00\%} & 62.61\% & \textbf{71.78\%} & \textbf{85.13\%} & \textbf{83.52\%} & 16 \\
    Runway Gen-3~\cite{runwaygen32024}            & 97.10\% & 96.62\% & 98.61\% & 99.23\% & 60.14\% & 63.34\% & 66.82\% & 84.11\% & 82.32\% & 256\\
    LaVie-2~\cite{wang2023lavie}           & 97.90\% & \textbf{98.45\%} & 98.76\% & 98.42\% & 31.11\% & \textbf{67.62\%} & 70.39\% & 83.24\% & 81.75\% & 61 \\
    Pika-1.0~\cite{PikaLabs2024}          & 96.94\% & 97.36\% & \textbf{99.74\%} & \textbf{99.50\%} & 47.50\% & 62.04\% & 61.87\% & 82.92\% & 80.69\% & 72\\
    VideoCrafter-2.0~\cite{chen2024videocrafter2}  & 96.85\% & 98.22\% & 98.41\% & 97.73\% & 42.50\% & 63.13\% & 67.22\% & 82.20\% & 80.44\% & 16 \\
    OpenSora V1.2~\cite{opensora}     & 96.75\% & 97.61\% & 99.53\% & 98.50\% & 42.39\% & 56.85\% & 63.34\% & 81.35\% & 79.76\% & 408\\
    \midrule
    LinGen & \textbf{98.30\%} & 97.60\% & 99.26\% & 98.58\% & 63.67\% & 60.55\% & 63.36\% & 83.77\% & 81.76\% & \textbf{1088}\\
    \bottomrule
  \end{tabular}
  \begin{tabular}{@{}l|>{\centering}p{0.97cm}c>{\centering}p{0.97cm}>{\centering}p{0.97cm}c>{\centering}p{0.97cm}>{\centering}p{0.97cm}>{\centering}p{0.97cm}>{\centering}p{0.97cm}|c@{}}
    \toprule
    \multirow{2}{*}{\textbf{Model}} & Object & Multiple  & Human  & \multirow{2}{*}{Color} & Spatial  & \multirow{2}{*}{Scene} & Appear.  & Temp.  & Overall  & \textbf{Semantic} \\
        & Class & Objects & Action &  & Relatio. & & Style & Style & Consist. & \textbf{Score} \\
    \midrule
    T2V-Turbo-v2~\cite{li2024t2vturbov2}       & \textbf{95.33\%} & 61.49\% & 96.20\% & 92.53\% & 43.32\% & \textbf{56.40\%} & 24.17\% & \textbf{27.06\%} & \textbf{28.26\%} & \textbf{77.12\%}  \\
    Runway Gen-3~\cite{runwaygen32024}             & 87.81\% & 53.64\% & 96.40\% & 80.90\% & 65.09\% & 54.57\% & 24.31\% & 24.71\% & 26.69\% & 75.17\% \\
    LaVie-2~\cite{wang2023lavie}            & 97.52\% & \textbf{64.88\%} & 96.40\% & 91.65\% & 38.68\% & 49.59\% & 25.09\% & 25.24\% & 27.39\% & 75.76\% \\
    Pika-1.0~\cite{PikaLabs2024}          & 88.72\% & 43.08\% & 86.20\% & 90.57\% & 61.03\% & 49.83\% & 22.26\% & 24.22\% & 25.94\% & 71.77\% \\
    VideoCrafter-2.0~\cite{chen2024videocrafter2}  & 92.55\% & 40.66\% & 95.00\% & \textbf{92.92\%} & 35.86\% & 55.29\% & \textbf{25.13\%} & 25.84\% & 28.23\% & 73.42\% \\
    OpenSora V1.2~\cite{opensora}     & 82.22\% & 51.83\% & 91.20\% & 90.08\% & \textbf{68.56\%} & 42.44\% & 23.95\% & 24.54\% & 26.85\% & 73.39\% \\
    \midrule
    LinGen & 90.98\% & 55.15\% & \textbf{97.50\%} & 83.95\% & 58.15\% & 53.51\% & 21.08\% & 24.29\% & 26.32\% & 73.73\% \\
    \bottomrule
  \end{tabular}
  \caption{Automatic evaluation of LinGen on \textbf{VBench-standard}. \textbf{Quality Score} measures the quality of generated videos and \textbf{Semantic Score} measures text-video alignment. \textbf{Total Score} represents their weighted sum. Higher values indicate better performance for all these metrics. }
  \label{tab:vbench_standard}
\end{table*}

\begin{table*}
  \centering
  \begin{tabular}{@{}l|cccccc|c@{}}
    \toprule
    \multirow{2}{*}{\textbf{Model}} & Subject  & Background   & Motion  & Aesthetic  & Imaging  & Dynamic  & \textbf{Quality}  \\
    & Consistency & Consistency  & Smoothness & Quality & Quality & Degree & \textbf{Score}  \\
    \midrule
    Sora~\cite{sora2024}               & 94.96\% & 95.84\% & 98.93\% & 60.30\% & 57.70\% & \textbf{69.30\%} & \textbf{79.69\%} \\
    Runway Gen-2~\cite{runwaygen32024}           & \textbf{97.61\%} & 97.61\% & \textbf{99.58\%} & \textbf{66.96\%} & 63.58\% & 18.89\% & 78.79\% \\
    Pika~\cite{PikaLabs2024}             & 96.76\% & \textbf{98.95\%} & 99.51\% & 63.15\% & 54.73\% & 37.22\% & 78.26\% \\
    VideoCrafter-1.0~\cite{chen2023videocrafter1}   & 95.10\% & 98.04\% & 95.67\% & 62.67\% & 61.99\% & 55.00\% & 78.14\% \\
    Show-1~\cite{zhang2024show}             & 95.53\% & 98.02\% & 98.24\% & 57.35\% & 59.75\% & 44.44\% & 77.50\% \\
    LaVie-Interpolation~\cite{wang2023lavie} & 92.00\% & 97.33\% & 97.82\% & 54.00\% & 59.78\% & 46.11\% & 75.86\% \\
    LaVie~\cite{wang2023lavie}              & 91.41\% & 97.47\% & 96.38\% & 54.94\% & 61.90\% & 49.72\% & 75.75\% \\
    ModelScope~\cite{wang2023modelscope}         & 89.87\% & 95.29\% & 95.79\% & 52.06\% & 58.57\% & 66.39\% & 74.91\% \\
    VideoCrafter-0.9~\cite{chen2023videocrafter1}   & 86.24\% & 92.88\% & 91.79\% & 44.41\% & 57.22\% & 89.72\% & 71.53\% \\
    CogVideo~\cite{hong2022cogvideo}           & 92.19\% & 96.20\% & 96.47\% & 38.18\% & 41.03\% & 42.22\% & 68.14\% \\
    \midrule
    LinGen              & 94.00\% & 96.08\% & 98.82\% & 57.86\% & \textbf{67.39\%} & 44.92\% & 78.59\% \\
    \bottomrule
  \end{tabular}
  \caption{\textbf{VBench-Custom} results based on customized prompts. \textbf{Quality Score} represents the weighted sum of these supported metrics.}
  \label{tab:vbench_custom}
\end{table*}

\begin{table*}
  \centering
  \begin{tabular}{@{}l|cccccc|c|c@{}}
    \toprule
    \multirow{2}{*}{\textbf{Model}} & Subject  & Background   & Motion  & Aesthetic  & Imaging  & Dynamic  & \textbf{Quality} & \textbf{Human Eval.} \\
    & Consistency & Consistency  & Smoothness & Quality & Quality & Degree & \textbf{Score} & \textbf{Win Rate} \\
    \midrule
    LinGen  @ 512p  & 94.00\% & 96.08\% & 98.82\% & 57.86\% & 67.39\% & 44.92\% & 78.59\% & 88.4\% \\
    LinGen @ 256p & 93.61\%	& 96.55\%	& 98.84\%	& 48.83\%	& 53.92\%	& 66.98\%	& 78.19\% & 11.6\% \\
    \bottomrule
  \end{tabular}
  \caption{VBench-Custom results of LinGen at different resolutions. Higher-resolution videos obtain a much higher win rate in human evaluation but only obtain a slightly higher VBench quality score. This indicates that VBench does not perfectly align with human preference.}
  \label{tab:vbench_custom_res}
\end{table*}

We provide a more complete VBench-Long leaderboard in Table~\ref{tab:vbench_long_comp}. We also evaluate LinGen on the standard VBench and compare it with other models on this leaderboard in Table~\ref{tab:vbench_standard}. Note that most models on this leaderboard can only generate very short videos (usually shorter than 5 seconds). VBench also provides the option to perform evaluations with customized prompts, although only some of the quality metrics are supported. We evaluate LinGen with Movie Gen Bench prompts~\cite{polyak2024moviegen} and compare it with other models on the VBench-Custom leaderboard in Table~\ref{tab:vbench_custom}.

The VBench results do not perfectly align with human preference. We find that Kling is more preferred in human evaluation than Runway Gen-3, but it obtains a lower VBench score. To further illustrate this point, as shown in Table~\ref{tab:vbench_custom_res}, when we evaluate our model at 256p and 512p resolutions on VBench-Custom, they obtained similar scores. However, 512p-generated videos have a much higher win rate than 256p-generated videos in human evaluation of video quality.

\subsection{Visual Examples and Human Evaluation}
\label{sec:app_viz}

\begin{figure}[t]
  \centering
   \includegraphics[width=\linewidth]{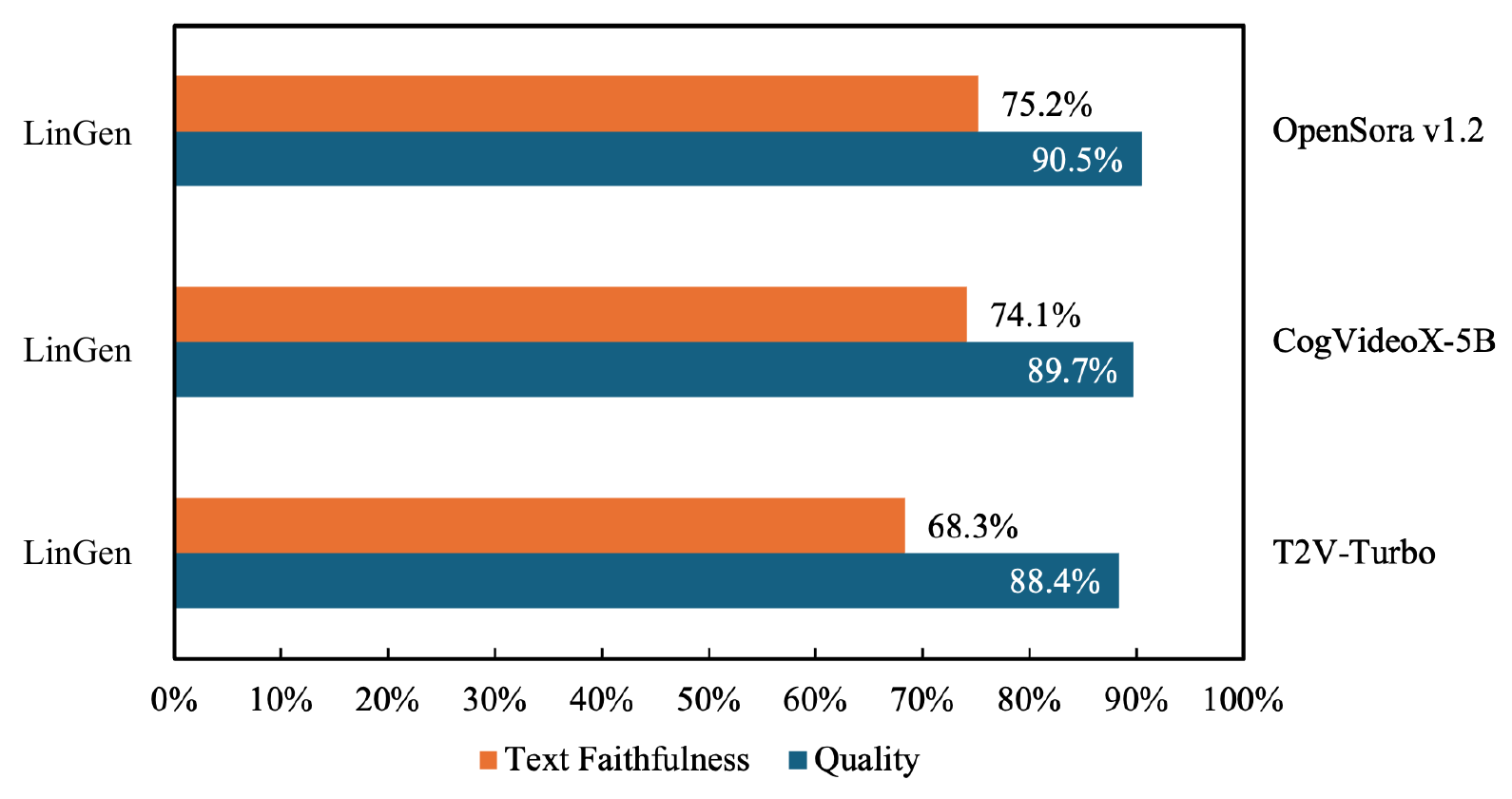}
   \caption{Win rates of human evaluation of quality and text-video alignment of videos generated by LinGen and typical open-source video generative models.}
   \label{fig:app_human}
\end{figure}

Given that the VBench results do not perfectly align with human preference, we provide more visual examples and human evaluation results to demonstrate the high quality of videos generated by LinGen in Fig.~\ref{fig:sotacomp} and Fig.~\ref{fig:app_human}, respectively. Fig.~\ref{fig:app_human} shows that LinGen outperforms typical open-source video generative models by a large margin.

\section{More Ablation Experiments}

We provide more visual examples of ablation experiments on the TESA block, RMS, review tokens, hybrid training, and quality-tuning in Fig.~\ref{fig:ablation_module} and Fig.~\ref{fig:ablation_tech}. This indicates that all of them contribute effectively to the consistency and high quality of the videos generated.

\section{Model Implementation Details}

In this section, we first provide more details of our model backbone in Sec.~\ref{sec:backbone}. Then, we compare Mamba and Mamba2 and present their technical details in Sec.~\ref{sec:mambacomp}. Finally, we give the details of our training recipe in Sec.~\ref{sec:recipe_app}.

\subsection{Backbone Details}
\label{sec:backbone}

LinGen learns a spatiotemporally compressed latent space using a Temporal AutoEncoder (TAE), designed similarly to the one in a prior work~\cite{polyak2024moviegen}. The TAE achieves a temporal compression rate of 8$\times$ and a spatial compression rate of 8$\times$8, followed by a 2$\times$2$\times$1 patchification. LinGen uses a factorized learnable positional embedding~\cite{dehghani2024patch} to enable arbitrary video size and length. We employ RMSNorm~\cite{zhang2019rmsnorm} and SwiGLU~\cite{shazeer2020swiglu} in LinGen, with adaptive layer normalization conditioned on the time step~\cite{peebles2023dit}.

\begin{figure}[t]
  \centering
   \includegraphics[width=\linewidth]{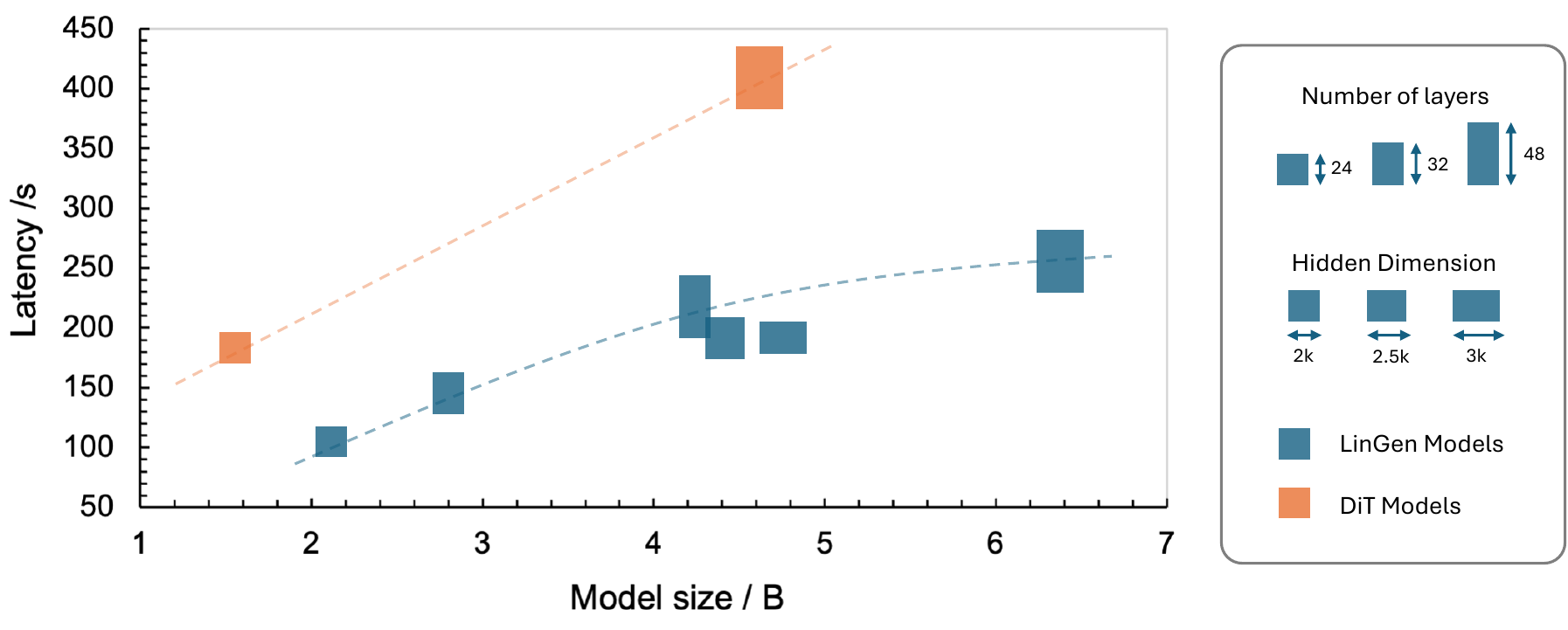}
   \caption{Latency of generating 512p 17s videos with different model designs. The latency of LinGen models scales more slowly with model size than self-attention-based standard DiT models. Note that we perform 100 inference steps to measure average latency. This is different from the default setting of 50 steps employed in our main paper.}
   \label{fig:archiexp}
\end{figure}

After completing architectural design exploration depicted in Fig.~\ref{fig:archiexp}, we employ 32 layers with 20 heads in each, with the dimension of embedding vectors being 2560.

\subsection{Mamba and Mamba2}
\label{sec:mambacomp}

SSMs have gained popularity in the field of natural language processing due to their high efficiency and strong performance in handling long sequences~\cite{gu2021ssm,gu2020hippo}. Mamba~\cite{gu2023mamba}, as a variant of SSM, enhances efficiency significantly by incorporating dynamic parameters into the SSM structure and developing algorithms optimized for better hardware compatibility. 
Early explorations~\cite{teng2024dimimg,yan2024diffussm} replaced the attention layers in diffusion models with SSMs, such as Mamba, to perform image generation, but these prototypes stayed relatively small. To unlock better efficiency at large scale, Mamba2~\cite{dao2024mamba2} unifies SSMs and the masked efficient attention by proposing a special SSM with an attention format (\ie, Structured State Space Duality). Mamba2 removes sequential linear projections that are used in Mamba and produces SSM parameters $A,B,C$ in parallel. The normalization layer in Mamba2 is the same as that in~\cite{shleifer2021normformer}. It improves stability. As mentioned in our main paper, the FLOPs cost of a bidirectional Mamba2 module is given by

\begin{equation}
    C_{\text{bimamba}} = (6+\frac{2}{d_h})ENd^2 + 4Nd_sd + O(Nd),
\end{equation}

\noindent
where $E$ is the expansion factor, $d$ is the dimension of token embedding vectors, $N$ is the number of tokens, $d_s$ is the hidden state size, and $d_h$ is the head dimension of Mamba2, whose default value is 64. $O(Nd)$ includes the FLOPs cost of 1D convolution and the SSM block in Mamba2:

\begin{equation}
    C_{\text{conv}} = 2EK(N+K-1)d
\end{equation}

\begin{equation}
    C_{\text{SSM}} = 4ENd_sd + 2ENd
\end{equation}

\noindent
where $K$ is the kernel size of 1D convolution. The above FLOPs should be doubled when the module is bidirectional.

Compared to Mamba, Mamba2 (1) has an attention format and thus benefits from existing efficient attention kernels, such as FlashAttention~\cite{dao2022flashattention} and xFormers~\cite{xFormers2022}, (2) supports much larger hidden state sizes with lower latency, and (3) has better support for tensor parallelism for upscaling of the model~\cite{waleffe2024empiricalmamba}.

Although Mamba2 compromises expressive power due to the simplification of the decay matrix in an SSM~\cite{dao2024mamba2}, it compensates for this using a much larger hidden state size. We set the hidden state size to 16 and 128 in LinGen w/ Mamba and LinGen w/ Mamba2, respectively, for both quality comparison and latency measurement, following their default values in the original design~\cite{dao2024mamba2}. 

\subsection{Training Recipe Details}
\label{sec:recipe_app}

In this Section, we introduce our progressive training recipe in Sec.~\ref{sec:progressive}. Then, we discuss our text-to-image and text-to-video hybrid training setting in Sec.~\ref{sec:hybrid}. We describe the details of our training datasets and quality-tuning design in Sec.~\ref{sec:dataset}.

\begin{table}
  \centering
  \begin{tabular}{@{}lccc@{}}
    \toprule
    Stage & \# Steps & Batch size & GPU days \\
    \midrule
    256p text-to-image         & 118k & 8192 & 1189 \\
    256p text-to-video, 17s    & 125k & 1024 & 1919 \\
    512p text-to-video, 17s    & 32k  & 512  & 2598 \\
    512p text-to-video, 34s    & 14k  & 512  & 2392 \\
    512p text-to-video, 68s    & 6k  & 256  & 1307 \\
    \bottomrule
  \end{tabular}
  \caption{The pre-training recipe of LVGen. The model was trained on Nvidia H100 GPUs.}
  \label{tab:recipe}
\end{table}

\subsubsection{Progressive Training Recipe}
\label{sec:progressive}

We use a progressive recipe to pre-train our LinGen-4B model. As shown in Table~\ref{tab:recipe}, we first pre-train our model on the text-to-image task at a 256p resolution, followed by text-to-video pre-training at progressively higher resolutions and longer video lengths. In this progressive training schedule, the token sequence length in the latent space gradually increases. 

\subsubsection{Hybrid Training}
\label{sec:hybrid}
In the text-to-video pre-training stages, we incorporate text-image pairs into the pre-training dataset and perform text-to-image and text-to-video joint training in practice. 
The sampling ratio of text-image pairs to text-video pairs is 1:100, which is very small, preventing this hybrid setting from reducing the motion of generated videos. We find such a hybrid training setting not only maintains the model's ability to generate images but also improves consistency of generated videos in some failure cases.

\subsubsection{Quality Tuning and Datasets}
\label{sec:dataset}

We use a progressive training schedule to train our DiT-4B and LinGen-4B models. (1) Text-to-image pre-training at 256p resolution. We use the licensed ShutterStock~\cite{shutterstock} image dataset, which includes 300M text-image pairs, to train our models. (2) Text-to-video pre-training at 256p and 512p resolutions to generate 17s videos. We use the licensed ShutterStock video dataset, which includes 24M text-video pairs, to train our models. (3) Text-to-video pre-training at 512p resolution to generate 34s and 68s videos. We select 2.5M videos that are longer than 30 seconds from the licensed ShutterStock video dataset to train our models. (4) Text-to-video pre-training at 512p resolution to generate 68s videos. We select 145K videos that are longer than 60s from the licensed ShutterStock video dataset to train our models. (5) Text-to-video quality tuning at 512p resolution. For the 17s video generation, we select 3K videos with extremely high quality and good motions from the ShutterStock and RawFilm~\cite{rawfilm} video dataset to fine-tune our model. For 68s video generation, we select 300 minute-length videos with high quality and good motions from the ShutterStock video dataset to fine-tune our model.

The way that we select high-quality videos is similar to that in prior works~\cite{dai2023emu,polyak2024moviegen}. We first filter videos via automatic metrics, including aesthetic score and motion score. Then, we balance the concepts in the set of videos, manually identify cinematic videos, and manually caption the videos.